\documentclass{article} 
\usepackage[preprint]{colm2026_conference}
\usepackage{adjustbox}

\definecolor{mydarkblue}{rgb}{0,0.08,0.45}

\usepackage{microtype}
\usepackage[colorlinks=true,
    citecolor=mydarkblue,
    filecolor=mydarkblue,
    urlcolor=mydarkblue,
    backref=page]{hyperref}
\usepackage{url}
\usepackage{booktabs}
\usepackage[utf8]{inputenc} 
\usepackage[T1]{fontenc}    
\usepackage{url}            
\usepackage{booktabs}       
\usepackage{amsfonts}       
\usepackage{nicefrac}       
\usepackage{microtype}      
\usepackage[table]{xcolor}         
\usepackage{amsmath}
\usepackage{graphicx}
\usepackage{bbm}
\usepackage{graphicx}
\usepackage{subcaption}  
\usepackage{float} 
\usepackage{multirow}
\usepackage{colortbl}
\usepackage{array}
\usepackage{makecell}
\usepackage{wrapfig}

\usepackage{enumitem}
\usepackage{lineno}
\definecolor{baselinebg}{HTML}{E8EAF6}
\definecolor{groupA}{HTML}{FAFAFA}
\definecolor{groupB}{HTML}{F0F0F0}
\definecolor{headerbg}{HTML}{37474F}
\definecolor{headertext}{HTML}{FFFFFF}
\definecolor{subheaderbg}{HTML}{78909C}
\definecolor{accent}{HTML}{1565C0}

\newcolumntype{L}{>{\raggedright\arraybackslash}p{1.4cm}}
\newcolumntype{M}{>{\centering\arraybackslash}p{0.85cm}}
\newcolumntype{R}{>{\centering\arraybackslash}p{1.5cm}}

\definecolor{darkblue}{rgb}{0, 0, 0.5}
\hypersetup{colorlinks=true, citecolor=darkblue, linkcolor=darkblue, urlcolor=darkblue}

\title{REAM: Merging Improves Pruning of Experts in LLMs}

\def\huggingface{\raisebox{-1.5pt}{\includegraphics[height=1.05em]{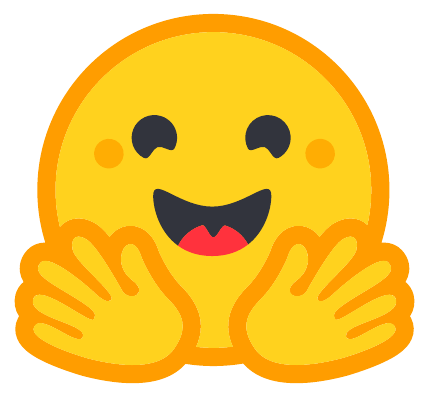}}}
\def\github{\raisebox{-1.5pt}{\includegraphics[height=1.05em]{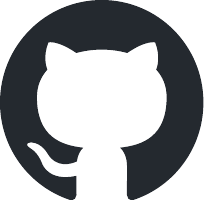}}}

\author{
{
Saurav Jha$^{1,2*}$ \:\:
Maryam Hashemzadeh$^{1,3}$ \:\:
Ali Saheb Pasand$^{1,4}$ \:\:
Ali Parviz$^{1}$}
\vspace{1mm}\\
{\textbf{\;\;Min-Joong Lee$^5$\quad\quad
Boris Knyazev$^{1,3,6*}$}} \vspace{2mm} \\
{$^{1}$Mila -- Quebec AI Institute \quad 
$^{2}$Polytechnique Montréal \quad 
$^{3}$Université de Montréal}\\
{$^{4}$McGill University \:
$^{5}$AI Center, Samsung, South Korea \:
$^{6}$Samsung AI Lab, Montreal}\\
{\;{\footnotesize Correspondence:\; \texttt{b.knyazev@samsung.com}
\quad\quad\quad $^*$equal contribution
}}\vspace{3mm}\\
{\github \: \textcolor{mydarkblue}{\footnotesize \url{https://github.com/SamsungSAILMontreal/ream}}}\vspace{1mm}\\
{\huggingface \: {\footnotesize \url{https://huggingface.co/collections/SamsungSAILMontreal/ream}}}
}

\begin{document}

\ifcolmsubmission
\linenumbers
\fi

\maketitle

\begin{abstract}
Mixture-of-Experts (MoE) large language models (LLMs) are among the top-performing architectures. The largest models, often with hundreds of billions of parameters, pose significant memory challenges for deployment. Traditional approaches to reduce memory requirements include \textit{weight pruning} and quantization. Motivated by the Router-weighted Expert Activation \textit{Pruning} (REAP) that prunes experts, we propose a novel method, Router-weighted Expert Activation \textit{Merging} (REAM). Instead of removing experts, REAM groups them and merges their weights, better preserving original performance. We evaluate REAM against REAP and other baselines across multiple MoE LLMs on diverse multiple-choice (MC) question answering and generative (GEN) benchmarks. 
Our results reveal a trade-off between MC and GEN performance that depends on the mix of calibration data.
By controlling the mix of general, math and coding data, we examine the Pareto frontier of this trade-off and show that REAM often outperforms the baselines and in many cases is comparable to the original uncompressed models.\looseness-1
\end{abstract}

\section{Introduction}

Mixture-of-Experts (MoE) layers replace a standard feed-forward block in a modern Transformer architecture \citep{vaswani2017attention} with a set of experts and a router that activates only a small subset of them for each token~\citep{jacobs1991adaptive, shazeer2017outrageously}. This conditional computation mechanism allows a model to grow dramatically in parameter count, while keeping the active per-token compute budget comparatively small. For modern LLMs whose performance benefits from scale, MoEs present a practical large-scale design architecture~\citep{jiang2024mixtral, liu2024deepseek, yang2025qwen3, team2025kimi}. For instance, Switch Transformers \citep{fedus2022switch} showed that sparse routing can push such models toward trillion-parameter scale without a commensurate increase in inference FLOPs. However, this efficiency comes with a fundamental trade-off. While MoEs reduce active computation, \textbf{all experts must still be stored, so they often trade FLOPs for memory and remain difficult to adapt in resource-constrained settings}.\looseness-1

A growing line of works suggests that the large parameter budget of MoEs is not used as effectively as intended because many experts become redundant~\citep{chi2022representation,Liu2023DiversifyingTM,limerge,jaiswal2025finding}.
\textbf{These  motivate the search for methods that remove the redundancy among similar experts without significantly sacrificing model performance.}
Inspired by traditional compression methods \citep{frantar2022gptq, lin2024awq}, MoE-based works address the above redundancy through two main directions: expert \textit{pruning} \citep{he2024demystifying,lasby2025reap} and \textit{merging} \citep{limerge,chen2025retrainingfree}. These two directions have certain trade-offs. On the one hand, merging preserves more information about all the original experts, but it depends critically on the quality of the grouping mechanism and can force suboptimal or functionally mismatched experts into the same group. On the other hand, pruning avoids the issues of grouping by dropping the original experts. In particular, \cite{lasby2025reap} proposed Router-weighted Expert Activation Pruning (REAP) that showed benefits of pruning compared to simple merging techniques. Despite their results, \textbf{the removal of experts may discard their useful knowledge, so REAP may not optimally balance the trade-offs between pruning and merging strategies}.\looseness-1

To better balance the trade-off between pruning and merging, we propose \textbf{Router-weighted Expert Activation Merging (REAM)} that preserves the knowledge of all experts, while effectively being similar to pruning due to our expert grouping and weighting approaches (Section~\ref{sec:method}). Our key contributions are as follows:
\begin{itemize}[leftmargin=*]
\item \textbf{Method:} We propose REAM, a unified expert compression framework with \textbf{four key components} to balance the trade-offs between merging and pruning in MoE models: \textbf{(1)}~an expert similarity metric that combines gate logit similarities with softmax-scaled expert output similarities, capturing both routing-level and representation-level redundancy; \textbf{(2)} a pseudo-pruning strategy that produces a few large groups and many singletons simultaneously; \textbf{(3)} enhanced weight alignment through a more informed cost matrix using both activation-based and weight-based costs; \textbf{(4)} a sequential merging procedure that recomputes forward-pass statistics after each layer is merged. 
\item 
\textbf{Performance:} We evaluate REAM under a 25\% and a 50\% expert reduction regime on Qwen3 and GLM4.5 MoE LLMs~\citep{qwen2.5-1m, yang2025qwen3,zeng2025glm} using eight multiple-choice (MC) benchmarks and six generative (GEN) reasoning and coding benchmarks.
We also examine the choice of calibration data by controlling the mixing ratio of general text, math and code data, which allows us to reveal an inherent trade-off between MC and GEN performance.
We examine the Pareto frontier of this trade-off and show that REAM often outperforms the baselines. \textbf{In the 25\% reduction regime, REAM performs comparably to, or only slightly below, the original uncompressed models.}
\end{itemize}

\section{Related Work} 
\label{sec:related_work}

\textbf{SMoE compression.}
In Sparse Mixture-of-Experts (SMoE, or simply MoE as referred to in this paper), the memory footprint and associated model-loading and communication overhead is tied to the total number of experts, which incurs significant deployment cost even though inference compute is sparse~\citep{jiang2025moecap}.
This has led to work on MoE efficiency span both \emph{system-level} methods that reduce serving overhead without changing the model itself \citep{xue2024moe, muzio2024seer, cai2025shortcut}, and \emph{model-level} methods that shrink the deployed model via compression techniques like quantization \citep{dong2025stbllm}, low-rank  decomposition \citep{yang2024moe, mi2026effective}, pruning \citep{jaiswal2025finding} or merging \citep{limerge}. 
Model-level compression methods are either \emph{static} \citep{chen2025retrainingfree, lasby2025reap}, where a one-shot transformation is applied at deployment time with no additional training, or \textit{dynamic} \citep{muqeethsoft, nguyen2025camex}, where training-time updates are made to the model parameters and the router to recover accuracy. Our work is along the static direction, which is more pragmatic than a dynamic one for real-world settings constrained by compute, data availability, privacy constraints, or deployment pipelines that require deterministic, reproducible model transformations.\looseness-1

\textbf{Expert pruning and merging.}
Expert reduction methods in MoEs mainly follow two paradigms: \emph{pruning} and \emph{merging}. Pruning removes redundant experts through routing \citep{Chen2022TaskSpecificEP,lu2024not,xie2024pruning,lasby2025reap} or search-based \citep{yang2024moe} saliency criteria. 
Compared to pruning, \emph{merging} combines similar experts in the weight-activation space \citep{li2022branchtrainmerge, chen2025retrainingfree,limerge, zhangdiversifying, he2024demystifying, chen2025retrainingfree} or shared-subspace representations \citep{gu2025delta,li2026sub}. 
After the grouping step, merging often aligns the parameters of experts~\citep{he2023merging, limerge, tran2025linear}, and then form a merged expert via interpolation or other approaches~\citep{miao2025mergemoe,nguyen2026expert}. 
Pruning and merging can be followed by additional compression of experts using singular value decomposition~\citep{limerge,li2026sub}, quantization~\citep{he2024demystifying}, or by post-compression adaptation to recover lost performance~\citep{muzio2024seer,huang2025discovering}. In our work, we focus only on the merging step and further compression or adaptation can be complementary to our approach.\looseness-1

While there are many strong expert pruning and merging methods,
we build on REAP~\citep{lasby2025reap} that achieved state-of-the-art performance in large-scale settings under 25\% and 50\% compression regimes. However, REAP removes experts potentially discarding important knowledge especially on the tasks outside of the calibration data domain. Moreover, REAP's advantage over merging is based on the assumption that merging methods tie gate weights and that gate logits are independent from the experts, thereby incurring an irreducible error in merging, which may not be true in practice.

\section{Background}

\paragraph{MoE layer.} 
An MoE layer replaces the feed-forward network in each Transformer \citep{vaswani2017attention} block with a set of $N$ expert networks $\{E_i\}_{i=1}^{N}$ and a learned router producing scores $g(\mathbf{x})=\mathbf{x} W_g \in \mathbb{R}^N$ that are dependent on the input token $\mathbf{x} \in X$. Gate logits are then converted to probabilities $\sigma(\mathbf{x}) = \text{Softmax}({g}(\mathbf{x}))$, so the MoE output is:
\begin{equation}
\label{eq:moe_out}
\mathbf{y}(\mathbf{x}) = \sum\nolimits_{i=1}^{N} \pi(\mathbf{x})_i \, E_i(\mathbf{x}),
\end{equation}
where $\pi(\mathbf{x}) = \text{Mask}\big(\sigma(\mathbf{x}), \text{top-}k\big) \in \mathbb{R}^N$ are the masked gate logits that are set to zero for the logits not in the top-$k$ values of $\sigma(\mathbf{x})$; top-$k$ is a constant that is much smaller than $N$, e.g., $N=128$ and top-$k=8$ in Qwen3 models~\citep{yang2025qwen3}.

\textbf{Expert saliency.} Central to both merging and pruning is the notion of expert saliency score $S_i$ that estimates the $i$-th expert's importance. For example, routing frequency \citep{jaiswal2025finding} counts how often expert $i$ is selected among the top-$k$ experts:
\begin{equation}
S_i^{\text{freq}}
\;=\;
\frac{1}{|X|}
\sum\nolimits_{\mathbf{x} \in X}
\mathbbm{1}\!\left[i \in \text{Top-}k\!\big(\sigma(\mathbf{x})\big)\right],
\label{eq:saliency_freq}
\end{equation}
where $\text{Top-}k(\cdot)$ returns the indices of the top-$k$ largest scores. Frequency is simple, but it assumes
that all active experts contribute equally to the output, so it can overvalue experts that are chosen with small router scores. REAP refines this by weighting selections by an estimate of contribution magnitude to the layer output~\citep{lasby2025reap}:
\begin{equation}
S_i^{\text{reap}}
\;=\;
\frac{1}{|\mathcal{X}_i|}
\sum\nolimits_{\mathbf{x} \in \mathcal{X}_i}
\pi(\mathbf{x})_i\,\big\|E_i(\mathbf{x})\big\|_2,
\label{eq:saliency_reap}
\end{equation}
where $\mathcal{X}_i \in X$ is the set of tokens where expert $i$ is active. This formulation better preserves MoE layer outputs and is leveraged in our approach.

\paragraph{Expert similarity.} Expert merging methods typically start by computing the similarity $\delta$ between experts $i$ and $j$, usually based on expert outputs~\citep{limerge, chen2025retrainingfree}: 
\begin{equation}
\label{eq:expert-sim-output}
\delta_E(i, j) = \frac{1}{|X|}\sum\nolimits_{\mathbf{x} \in X} \text{sim}(E_i(\mathbf{x}), E_j(\mathbf{x})),
\end{equation}
where $\text{sim}(\cdot,\cdot)$ is a similarity metric, such as cosine similarity.
Alternatively, the similarity $\delta$ can be computed based on \textbf{gate logits} \citep{he2024demystifying}:
\begin{equation}
\label{eq:expert-sim-gate}
\delta_g(i, j) = \text{sim}\big(\bigl[g(\mathbf{x}_1)_i,\ldots,g(\mathbf{x}_{|X|})_i\bigr], \bigl[g(\mathbf{x}_1)_j,\ldots,g(\mathbf{x}_{|X|})_j\bigr]\big),
\end{equation}
where $g(\mathbf{x}_j)_i \in \mathbb{R}$ are the gate logits of expert $i$ for token $\mathbf{x}_j$ of the calibration data $X$.

\paragraph{Expert grouping and merging.} The second step of merging concerns grouping of similar experts. \cite{limerge} introduced a simple grouping method, in which first the experts with highest $S_i^{\text{freq}}$ are chosen as the group centroids. Then all other experts are assigned based on the expert similarity in Eq. \eqref{eq:expert-sim-output} or \eqref{eq:expert-sim-gate}. This procedure does not explicitly control the size $C$ of resulting groups. 
Expert merging is then done as the weighted average per group:
\begin{equation}
\label{eq:expert-avg}
\mathbf{W}_{\text{merged}} = \frac{\sum_{i=1}^C S_i^{\text{freq}}\mathbf{W}_i }{\sum_{i=1}^C S_i^{\text{freq}}},
\end{equation}
where $\mathbf{W}_i$ are expert $i$'s weight matrices with neuron permutation alignment~\citep{ainsworth2022git} applied w.r.t. the dominant (centroid) expert.

\paragraph{Gate weights.}
After obtaining a reduced set of experts, pruning methods typically remove the rows of the gate weights $W_g$ corresponding to the dropped experts as in REAP~\citep{lasby2025reap}. In contrast, merging methods keep gate weights as is and sum the gate logits per group, which can result in an irreducible error as shown by \cite{lasby2025reap} and discussed in Section~\ref{sec:related_work}. In our work, we follow REAP and remove the rows of the gate weights that are not corresponding to centroid experts.

\section{Router-weighted Expert Activation Merging}
\label{sec:method}

\paragraph{Aggregated expert similarity.}
We compute expert similarity as the sum of two similarities:
\begin{equation}
    \label{eq:expert-sim}
    \delta_{\text{REAM}}(i, j) = \delta_g(i, j) + \tilde{\delta}_E(i, j),
\end{equation}
where $\delta_g(i, j)$ is computed as in Eq.~\eqref{eq:expert-sim-gate} and our \textbf{gated expert similarity} $\tilde{\delta}_E(i, j)$ is computed based on Eq.~\eqref{eq:expert-sim-output}:
\begin{equation}
\label{eq:gated-expert-sim}
\tilde{\delta}_E(i, j) = \frac{1}{|X|}\sum\nolimits_{\mathbf{x} \in X} \text{sim}(\sigma(\mathbf{x})_i E_i(\mathbf{x}), \sigma(\mathbf{x})_j E_j(\mathbf{x})),
\end{equation}
where we use gated expert outputs $\sigma(\mathbf{x})_i E_i(\mathbf{x})$, which matches closely the computation of the MoE output in Eq.~\eqref{eq:moe_out}. It ensures that expert outputs are modulated by the gate, making the similarity metric aware of expert specialization.

\paragraph{Pseudo-pruning.} 
\label{sec:pseudo-prune}
Given REAP saliency scores ${S_i^{\text{reap}}}$ computed over a calibration set $X$, we group the $N$ experts into $N' < N$ clusters via a greedy pseudo-pruning procedure. Here, we follow \cite{limerge} and for each layer $\ell$, we designate the $N'$ experts with the highest saliency as the cluster centroids $\mathbf{C}_\ell = \{c_1, \ldots, c_{N'}\}$, but we sort them in decreasing order of saliency. Then, starting from $c_1$, we greedily assign to it up to $C$ unassigned non-centroid experts $E_j$ that are most similar to $c_1$ based on $\delta_{\text{REAM}}(c_1, j)$ in Eq.~\eqref{eq:expert-sim}.

\begin{figure}[t]
    \centering
\begin{subfigure}[t]{0.48\textwidth}
        \centering
        \includegraphics[width=0.75\textwidth]{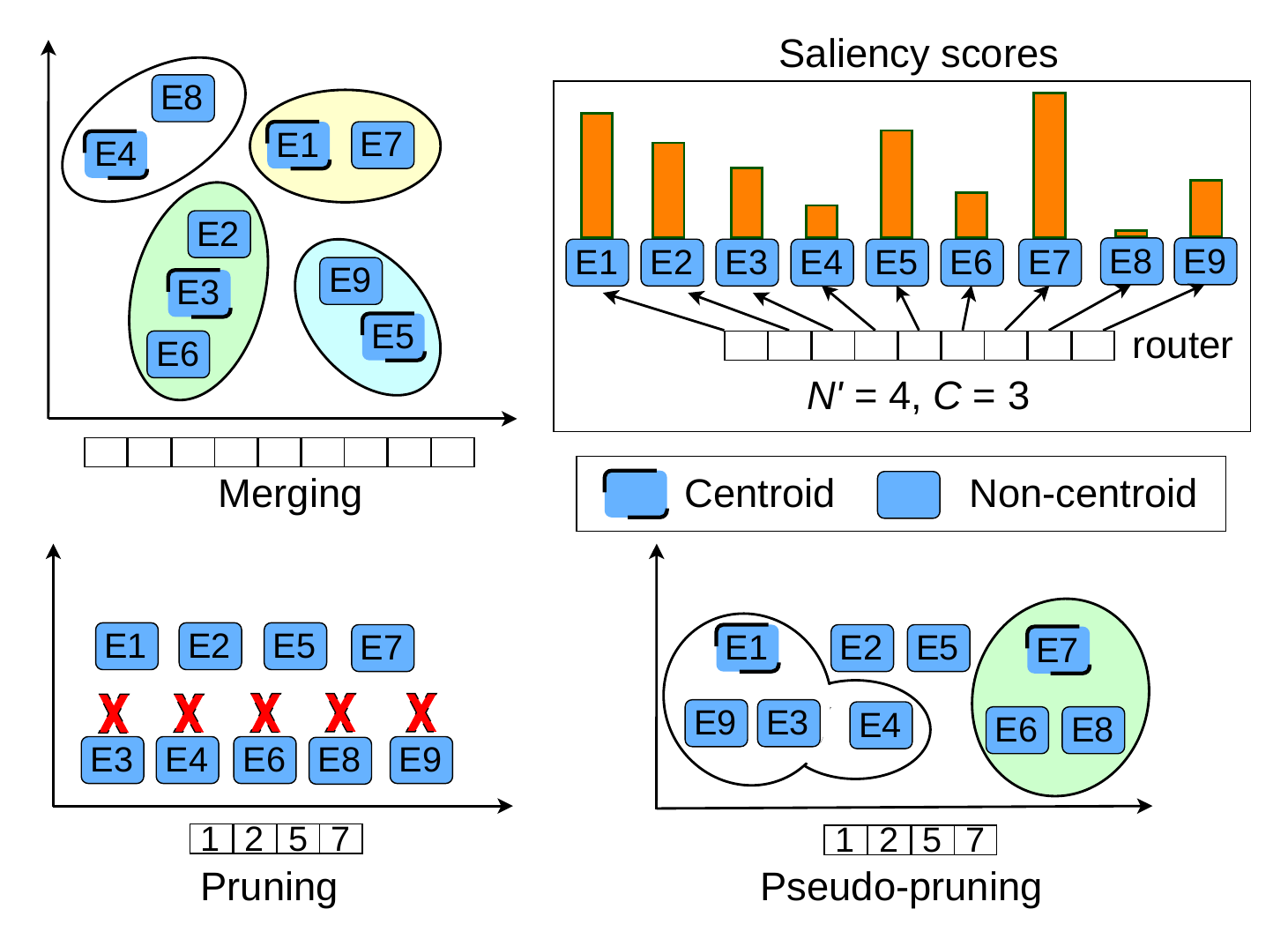}
        \caption{Merging vs. Pruning vs. Pseudo-pruning}
        \label{fig:pseudo_prune}
            \end{subfigure}    
        \hfill
    \begin{subfigure}[t]{0.51\textwidth}
        \centering
        \includegraphics[width=\textwidth]{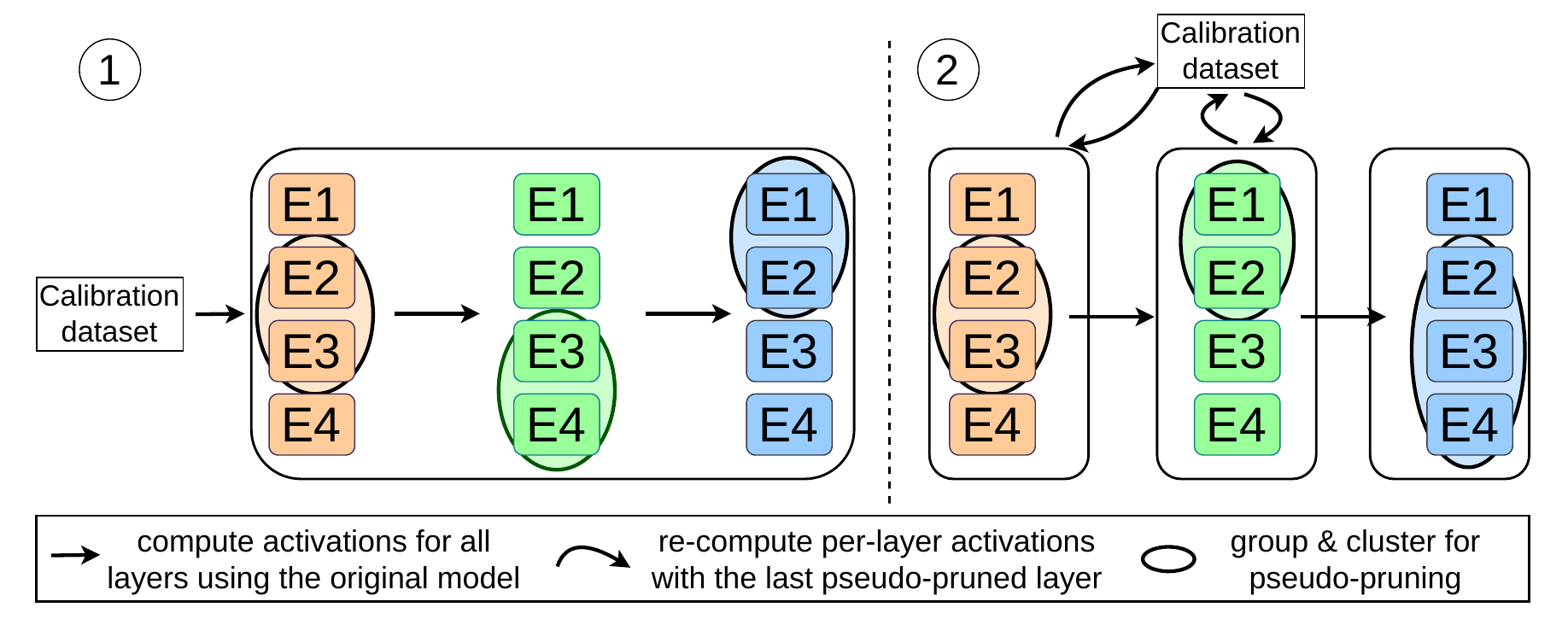}
        \caption{Sequential merging}
        \label{fig:sequential_merge}
    \end{subfigure}
    \caption{\textbf{Illustration of REAM components:} a) Comparison of expert compression strategies reducing $N{=}9$ experts to $N'{=}4$.  HC-SMoE merging~\citep{chen2025retrainingfree} clusters all experts by output similarity regardless of saliency (e.g., E1 and E7  grouped together). Pruning retains the top-4 salient experts unchanged and discards the rest.
    Our REAM's pseudo-pruning selects the top-4 experts as protected centroids and absorbs remaining experts into their nearest centroid via saliency-weighted merging, leaving other groups as singletons. 
    b)~Compared to baseline pruning and merging methods \raisebox{.5pt}{\textcircled{\raisebox{-.9pt} {1}}} that collect the activations from the original uncompressed model for all layers at once, REAM \raisebox{.5pt}{\textcircled{\raisebox{-.9pt} {2}}} recomputes the per-layer activations after merging each MoE layer before processing the next layer.}
    \label{fig:merge_vs_pseudoprune}
\end{figure}

Since typically $N - N' \ll N' \cdot C$ (e.g., $N'$ is 25\% smaller than $N$), the set of non-centroid experts is far smaller than the total absorption capacity of all centroids, so most centroids receive no assignments and form singleton groups that pass through unchanged. Accordingly, we call our grouping method  \textit{pseudo-pruning}. Unlike merging methods that tend to cluster experts into many medium-sized groups, pseudo-pruning results in a few large groups while many singletons are left intact  (Fig.~\ref{fig:pseudo_prune}).

\paragraph{Activation and weight permutation alignment.}
\label{sec:permutation}
In the expert merging step of Eq.~\eqref{eq:expert-avg}, the weights need to be aligned before computing their weighted average. For example, \cite{limerge} used the Hungarian algorithm with the cost matrix $\mathcal{C}_{\text{wt}}$ computed based on the distances between the weights of the centroid expert $c_i$ and expert $j$.
To improve the alignment, we introduce a combined cost matrix $\mathcal{C}_{\langle c_i, j\rangle} = \mathcal{C}_{\text{act}} + \mathcal{C}_{\text{wt}} \in \mathbb{R}^{d \times d}$. 
Here, $[\mathcal{C}_{\text{act}}]^{pq} = \|\bar{\mathbf{H}}_{c_i}^{p} - \bar{\mathbf{H}}_{j}^{q}\|_2$ is the distance between the normalized calibration-token activation vectors of the $p$-th and the $q$-th neurons across the two experts, and  $[\mathcal{C}_{\text{wt}}]^{pq} = \|\mathbf{W}_{c_i}^{p} - \mathbf{W}_{j}^{q}\|_2$ is the distance between their weight matrices $\mathbf{W}_i^{(p)}$. Thus, $\mathcal{C}_{\langle c_i, j\rangle}$ combines data-driven signal with a data-independent one so that a matched neuron pair must be consistent in both activation and weight space. The optimal permutation is then applied to reorder the weights of expert $j$.
Using data-based cost alone to find the optimal permutation can be noisy since two neurons might happen to produce similar activations on the calibration batch by coincidence, even if their weights are very different. On the contrary,  weight-based cost alone ignores how the model actually uses each neuron in practice -- two neurons with similar weights but very different activation patterns (due to how inputs distribute) are still suboptimal to merge. The combined cost matrix balances between both ends.

\paragraph{Sequential merging.}
\label{sec:sequential}
Prior expert pruning and merging methods run a single forward pass through the original, unmodified model to collect per-layer statistics. The pre-collected statistics are then used to compress all layers independently. However, once the experts in layer $\ell$ are compressed, its modified outputs render the statistics for the subsequent layers as stale. Instead, we propose 
 updating the model outputs to reflect the currently merged layers. As shown in Fig. \ref{fig:sequential_merge},  after merging layer $\ell$, a second forward pass is run through this layer to recompute its activations to be used by the subsequent layer $\ell+1$. This ensures that each layer's statistics reflect the actual input it will receive at inference time. 
 Since sequential merging requires computing the forward pass through a given MoE layer twice, a genuine concern remains its computational  overhead compared to non-sequential merging. However, in practice, we find it to be reasonably fast. For  Qwen3-30B-A3B-Instruct-2507 \citep{yang2025qwen3}, non-sequential merging takes $\approx$1 hour, while our sequential variant takes $\approx$1.5 hours, with $\approx$30 GB of VRAM in both cases.
 Given that merging is done only once for a given model, the effectiveness of this procedure  usually carries more significance than the efficiency.\looseness-1

\section{Experiments}
\label{sec:experiments}

\paragraph{Setup.} We follow evaluation in REAP~\citep{lasby2025reap} and evaluate all methods without any fine-tuning after compression. For our testbed, we focus primarily on Qwen3-30B-A3B-Instruct-2507 \citep{yang2025qwen3}, a 30B-parameter MoE model with $N=$128 experts per layer, of which top-$k$=8 are active per token. We  additionally validate on the larger Qwen3-Coder-Next and Qwen3-Next-80B-A3B-Instruct~\citep{cao2026qwen3}, both  80B-parameter models with 512 experts per layer, and on GLM-4.5-Air \citep{zeng2025glm}, a 106B-parameter model with 128 experts per layer. We compress models by merging 25\% or 50\% of the experts per layer, e.g., reducing from 128 to 96 or 64 experts, respectively. 

\paragraph{Calibration dataset.} For calibration, we collect router logits and expert activations on a mixture of three datasets with 3072 sequences of 512 tokens each —  C4~\citep{2019t5} for general language understanding, NuminaMath~\citep{numina_math_datasets} for mathematical reasoning, and The-Stack-Smol~\citep{Kocetkov2022TheStack} for code generation. To study the sensitivity of merging decisions $\,\text{w.r.t.}\,$ the calibration distribution, we experiment with ten different mixing ratios across these three sources, ranging from math-heavy (0.0:0.7:0.3) to code-heavy (0.1:0.1:0.8) configurations (see Table~\ref{tab:ratios} for the full table of ratios).  

\paragraph{Evaluation.} Compressed models are evaluated on two benchmark suites (see Section~\ref{app:mc_tasks} for details). The first consists of 8 multiple-choice (MC) tasks following prior work~\citep{chen2025retrainingfree,lasby2025reap}. 
The second consists of 6 generative (GEN) tasks: IFEval \citep{zhou2023instructionfollowingevaluationlargelanguage}, AIME25 \citep{aime25}, GSM8K \citep{cobbe2021gsm8k}, HumanEval \citep{chen2021evaluating}, GPQA-Diamond \citep{rein2024gpqa}, and LiveCodeBench \citep{jain2025livecodebench}.
We report the mean score within each suite. Since generative tasks are typically more practically relevant and challenging, we present our key results on the GEN suite (Tables \ref{tab:main_results}, \ref{tab:analysis4}).\looseness-1

\paragraph{Baselines.} We compare REAM against two expert pruning baselines: frequency-based (Freq) and REAP~\citep{lasby2025reap}. HC-SMoE~\citep{chen2025retrainingfree} is used as a merging baseline with average linkage clustering and activation-based permutation alignment.
The only hyperparameter of REAM is group size $C$ of pseudo-pruning (Section~\ref{sec:pseudo-prune}), which is fixed to 16 or 32 depending on the number of experts (Section~\ref{app:hyperparam}).

 \subsection{Main Results}
 \label{sec:main_results}
 
\begin{figure}[t]
  \centering  \includegraphics[width=0.7\textwidth]{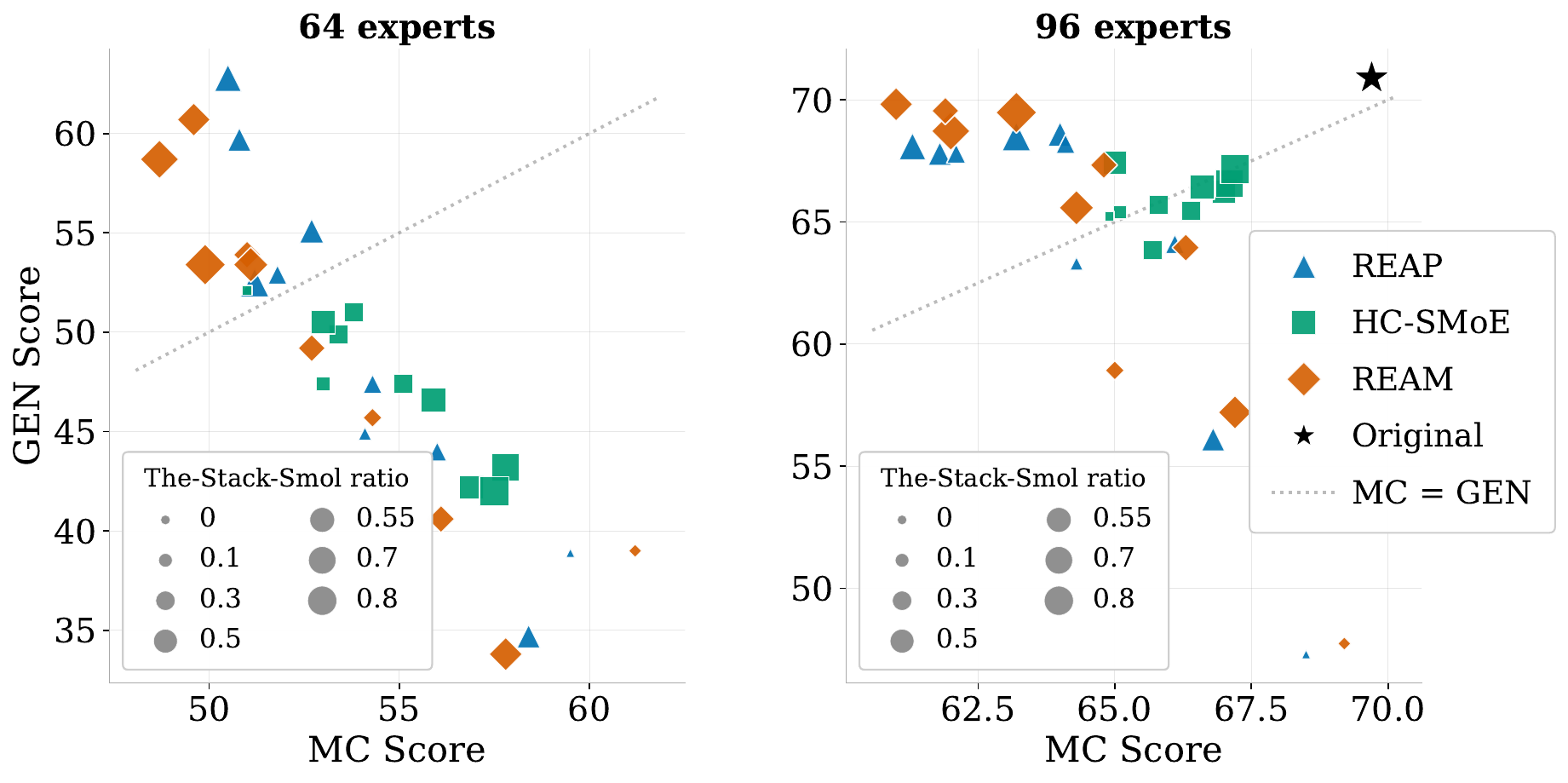}
  \caption{\textbf{Discriminative (MC) vs. Generative (GEN) trade-off depending on the calibration data mixture:} benchmark scores with 64 (\textbf{left}) and 96 (\textbf{right}) experts for REAP, HC-SMoE, and REAM across ten mixing ratios of the calibration data with Qwen3-30B-A3B-Instruct-2507. The marker sizes are proportional to the \textbf{The-Stack-Smol} share of the mixture.\looseness-1
  }
  \label{fig:main_results}
\end{figure}

 \paragraph{MC vs GEN results.}
We first compare REAM to baselines at 64 and 96 experts obtained with ten mixing ratios of the calibration dataset on both GEN and MC benchmarks (Fig. \ref{fig:main_results}).
We leave the detailed results across the mixing ratios of C4:Math:Code in Fig. \ref{fig:main_results_app} and Tables \ref{tab:gen_tasks_64}-\ref{tab:gen_tasks_96}.  Given their reliance on expert saliencies for compression, we observe the performance of Freq, REAP, and REAM to strongly depend on the calibration composition, but not for HC-SMoE. For Freq and REAP, calibrating without any code data (Code $=0$ corresponding to the smallest markers in Fig. \ref{fig:main_results}) is catastrophic for code-generation tasks, with HumanEval and LiveCodeBench scores collapsing to near zero despite strong math performance, \textit{i.e.,} a gap of over 40 points compared to the best configuration (Table~\ref{tab:gen_tasks_96}). 

 Similarly to Freq and REAP, REAM is also sensitive to the mixing ratio where its best ratio ($0{:}0.5{:}0.5$) achieves a GEN average of 69.8, within 1.1 points of the uncompressed 128-expert baseline (70.9), while its worst ratio of $0.5{:}0.5{:}0$ yields 47.7 (Table~\ref{tab:gen_tasks_64}). By contrast, HC-SMoE's best and worst averages span only 3.5 points (67.4 vs.\ 63.9), suggesting its saliency-independent clustering is robust to, but also unable to benefit from, task-aligned calibration. Overall, well-chosen data mixtures help REAM consistently outperform all baselines, with REAP standing second, and HC-SMoE and Freq being roughly tied (Table~\ref{tab:main_results}).

 \begin{table*}[thbp]
\centering
\footnotesize
\caption{Results at 96 experts on Qwen3-30B-A3B-Instruct-2507.
Each method uses the calibration mixture achieving its best GEN score;
\textbf{bold} is the best among compressed models.\looseness-1}
\label{tab:main_results}
\setlength{\tabcolsep}{4pt}
\renewcommand{\arraystretch}{1.08}
\rowcolors{3}{gray!6}{white}
\begin{tabular}{l c c|c c c c c c|c}
\toprule
Method & $N$ & C4:Math:Code & IFEval & AIME25 & GSM8K & GPQA & HumanEval & LCB & GEN \\
\midrule
Original & 128 & --            & 90.4 & 56.7 & 89.3 & 47.0 & 93.3 & 48.6 & 70.9 \\
\midrule
Freq     &  96 & 0:0.3:0.7     & 87.8 & $\mathbf{60.0}$ & 82.9 & 36.9 & 93.9 & 44.0 & 67.6 \\
HC-SMoE  &  96 & 0.5:0:0.5     & 88.2 & $\mathbf{60.0}$ & 84.7 & 34.3 & 91.5 & 45.9 & 67.4 \\
REAP     &  96 & 0.2:0.25:0.55 & 89.6 & 50.0 & \textbf{87.9} & \textbf{39.4} & \textbf{94.5} & 50.3 & 68.6 \\
\rowcolor{blue!8}
REAM     &  96 & 0:0.5:0.5     & \textbf{89.9} & $\mathbf{60.0}$ & 86.3 & 38.4 & 93.3 & $\mathbf{51.0}$ & \textbf{69.8} \\
\bottomrule
\end{tabular}
\end{table*}

\begin{figure}[t]
    \centering
     \begin{subfigure}[t]{0.55\textwidth}
        \centering
       \includegraphics[width=\textwidth]{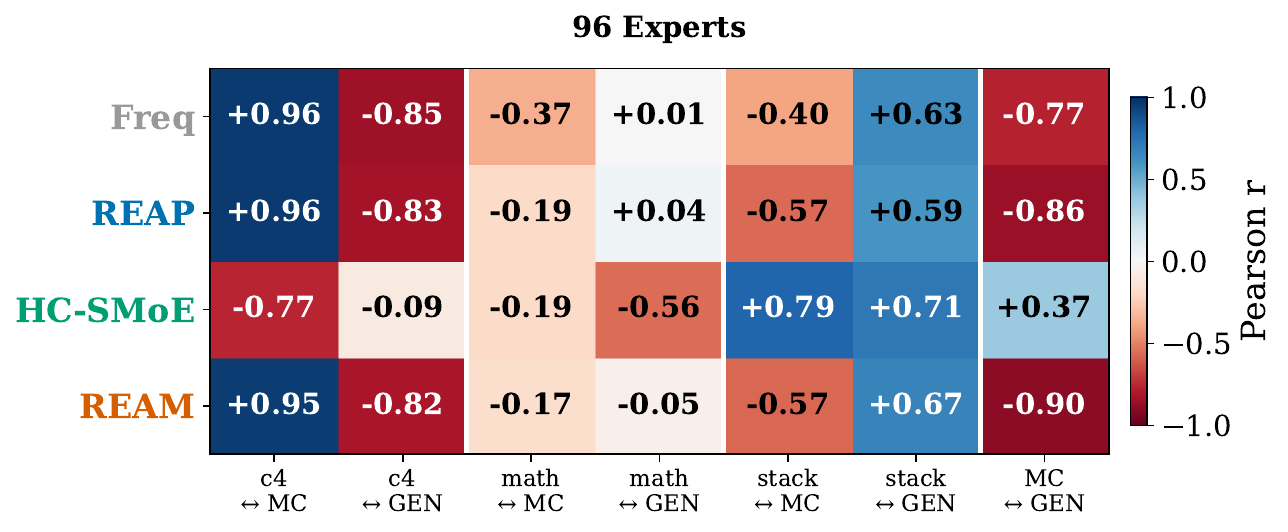}
        \caption{Correlation $r$ across methods and domains}
        \label{fig:correlations_96}
    \end{subfigure}
    \hfill
\begin{subfigure}[t]{0.44\textwidth}
        \centering
        \includegraphics[width=0.8\textwidth]{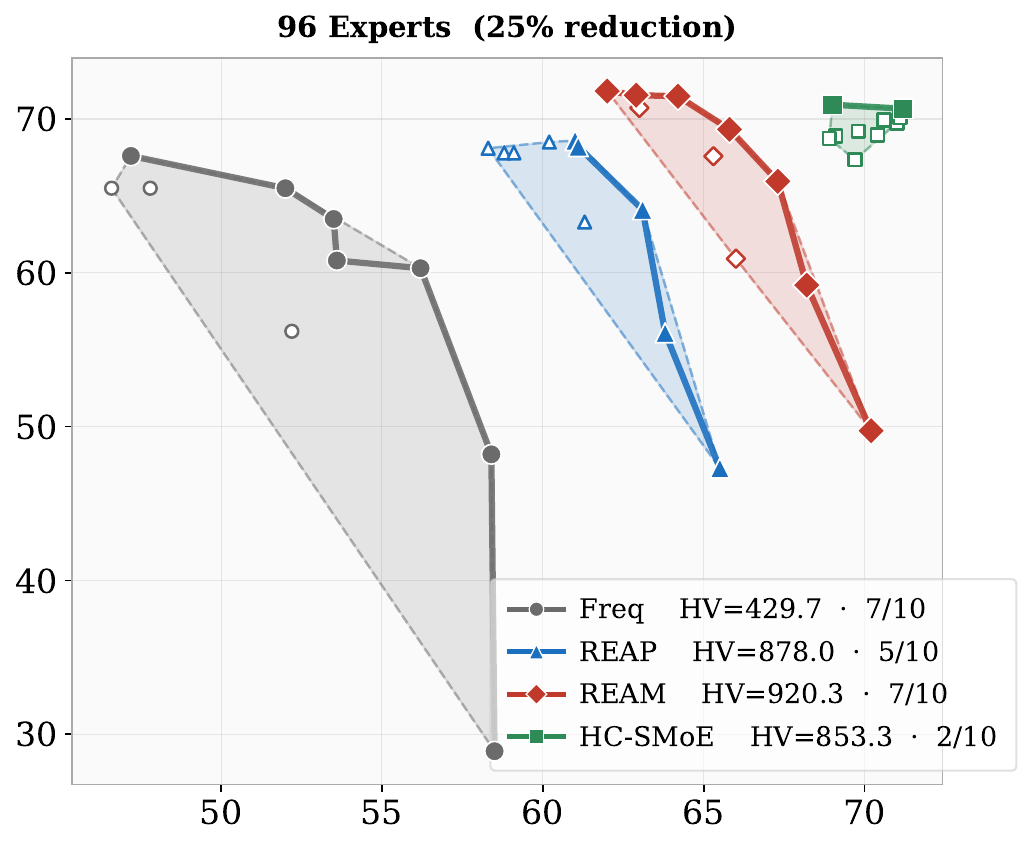}
        \caption{Pareto frontiers per method}
        \label{fig:pareto_96}
            \end{subfigure}       
    \caption{\textbf{Additional analyses for 96 experts:} a) Pearson correlation $r$ between calibration datasets (C4, Math, Code) and MC/GEN scores, and between MC and GEN scores themselves, for each merging method. b) Pareto frontiers where each point is one of 10 calibration mixtures. Filled markers denote Pareto-optimal configurations not simultaneously dominated on MC and GEN by any other mixture of the same method, and hollow markers denote dominated ones. The hypervolume (HV) measures the area of the MC$\times$GEN plane dominated by each method's frontier relative to a shared reference point, quantifying its overall performance ceiling.
    Per-method offsets are applied for better visibility.}
    \label{fig:task_sp_studies}
\end{figure}

\paragraph{Calibration data vs. performance correlation.} To understand the systematic structure underlying the calibration data mixtures, we further analyze the performance correlations $r$ for different methods on the 96-expert setting. Fig. \ref{fig:correlations_96} shows that for Freq, REAP, and REAM, the proportion of C4 data in the calibration mixture is strongly positively correlated with MC scores ($r \geq 0.95$) yet strongly negatively correlated with GEN scores ($r \leq -0.82$), indicating a fundamental MC--GEN trade-off driven by general-domain calibration. Conversely, Code proportion is consistently positively correlated with GEN ($r \geq 0.59$) while negatively correlated with MC ($r \leq -0.40$), and math proportion has negligible correlation with either suite. The strong negative MC--GEN correlation for these three methods shows that no single calibration dataset simultaneously maximizes both  performances.  HC-SMoE shows an exception to this trend. While its C4--MC correlation is strongly negative, its stack--MC and MC--GEN correlations are positive. Such counterintuitive behavior can be attributed to HC-SMoE's grouping decisions being largely invariant to what calibration data is provided. We provide further analysis and discussion in Section~\ref{app:correlations_96_detailed}.

\subsection{Pareto Analysis of MC vs GEN}

\paragraph{Setup.}
A real-world deployment scenario of a compressed MoE is often concerned with the best-case comparison across methods at equal performance levels, \textit{e.g.}, \textit{while preserving an MC score of 65, what is the best GEN any calibration ratio can achieve for REAM vs. HC-SMoE?} Hence, we study the sensitivity of each compression method to the choice of calibration mixture by examining each method's configurations in the joint MC$\times$GEN space. Here, each of the 10 mixing ratios yields one point per method. The convex hull enclosing all 10 points gives us the Pareto frontier, \textit{i.e.}, the subset of configurations that are not simultaneously dominated on both metrics by any other configuration of the same method. Lastly, to quantify how much of the MC$\times$GEN space each method's frontier occupies, we compute the \emph{hypervolume} (HV) indicator, \textit{i.e.}, the area of the MC$\times$GEN plane dominated by the Pareto frontier relative to a fixed reference point set one unit below the global minimum on each axis. A larger HV means the method can achieve better MC--GEN trade-offs across a wider range of calibration preferences. Together with the fraction of Pareto-optimal configurations $n/10$, which measures how many of the 10 ratios lie on the frontier, we characterize both the \emph{performance ceiling} and the \emph{calibration robustness} of each method.

\paragraph{Results.}
At 25\% compression, HC-SMoE has the lowest n/10 ($=2/10$) and low HV (853.3), meaning that nearly all its configurations are clustered in a tight band regardless of whether calibration data is majorly text, math, or code (Fig. \ref{fig:pareto_96}). While this was a design choice in HC-SMoE~\citep{chen2025retrainingfree}, our analysis reaffirms that HC-SMoE's performance envelope is narrow and that calibration selection offers little leverage. Freq shows the opposite failure mode — a high n/10 ($7/10$) driven by a wide scatter of configurations across the MC$\times$GEN plane, yet the lowest HV (429.7) of all methods. 
REAP achieves a higher HV (878.0) with a moderate n/10 ($5/10$), thus tracing a clearer MC--GEN trade-off curve that shifts predictably with calibration mixtures. However, its frontier saturates in the high-GEN region where code-heavy ratios dominate. Our REAM attains both the highest HV (920.3) and the highest n/10 ($7/10$). This shows that for virtually any MC floor, there exists a calibration mixture under which REAM's frontier dominates all other methods on GEN, confirming that its advantage is not confined to a single lucky ratio but holds broadly across the calibration space. Fig. \ref{fig:pareto_64} in Appendix further shows a similar analysis at 64 experts.

\subsection{Larger Models}  

\paragraph{Setup.}
We assess the effectiveness of REAM on two variants of Qwen3-Next 
 with a larger set of 512 experts and 80B parameters: Qwen3-Next-80B-A3B-Instruct \citep{yang2025qwen3} and Qwen3-Coder-Next \citep{cao2026qwen3}, and on GLM-4.5-Air \citep{zeng2025glm} with 128 experts and 106B parameters. 
 These models were evaluated without any additional tuning of REAM or baselines (other than fixing $C$ to 32 or 16, Section~\ref{app:hyperparam}). 
 Since performing merging and evaluation for all the mixing ratios is expensive, we fix the mixture at a code heavy ratio of \,0\,:\,0.3\,:\,0.7 to favor the overall GEN score following our analysis in Section~\ref{sec:main_results}. 
 Additional mixing ratios and ablations for Qwen3-Coder-Next are reported in Table~\ref{tab:qwen3-coder-ablations}. 
 
 \paragraph{Results.}
We show that REAM matches the GEN score of the uncompressed Qwen3-Coder-Next at $25\%$ compression, thus demonstrating near-lossless compression on a strong code model (Table~\ref{tab:analysis4}).
Moreover, REAM consistently outperforms REAP on GEN across all the three models.
On several tasks (IFEval, AIME25 and GSM8K), REAM often recovers the full original score while REAP lags behind. 
Similar to Table \ref{tab:main_results}, GPQA remains the most sensitive task where both the methods show notable drops. Further ablations of REAM on Qwen3-Coder-Next (Table \ref{tab:qwen3-coder-ablations}) show trends  similar to that of Qwen3-30B-A3B-Instruct-2507. Here, AIME25 is highly sensitive to the overall calibration mix while both GSM8K and HumanEval are boosted by code-heavy calibration with REAM to the point of surpassing the original uncompressed model.\looseness-1

\begin{table}[t]
\centering
\scriptsize
\setlength{\tabcolsep}{5pt}
\renewcommand{\arraystretch}{1.08}
\caption{GEN benchmark results on additional models with a 25\% expert reduction:
         512\,$\to$\,384 experts for Qwen3-Next-80B-A3B-Instruct and Qwen3-Coder-Next, and 128\,$\to$\,96 for GLM-4.5-Air. The calibration mixture is fixed at C4\,:\,Math\,:\,Code\,=\,0\,:\,0.3\,:\,0.7 to favor GEN tasks;
         \textbf{bold} is the best among compressed models.
         }
\label{tab:analysis4}
\begin{tabular}{llc|cccccc|c}
\toprule
\textbf{Model} & \textbf{Method} & $N$
  & \textbf{IFEval} & \textbf{AIME25}
  & \textbf{GSM8K} & \textbf{GPQA} & \textbf{HumanEval} & \textbf{LCB} & \textbf{GEN} \\
\midrule
\multirow{3}{*}{Qwen3-80B-A3B}
  & \cellcolor{black!4}Original  & \cellcolor{black!4}512 & \cellcolor{black!4}{93.4} & \cellcolor{black!4}{80.0} & \cellcolor{black!4}{78.6} & \cellcolor{black!4}{47.0} & \cellcolor{black!4}{95.1} & \cellcolor{black!4}43.2          & \cellcolor{black!4}72.9          \\
  & REAP      & 384 & 92.8          & 66.7          & 77.7          & 42.4          & \textbf{94.5}          & 43.6          & 69.6          \\
  & \cellcolor{blue!8}REAM      & \cellcolor{blue!8}384 & \cellcolor{blue!8}\textbf{93.4} & \cellcolor{blue!8}\textbf{73.3}          & \cellcolor{blue!8}\textbf{78.1}          & \cellcolor{blue!8}\textbf{46.5}          & \cellcolor{blue!8}93.9          & \cellcolor{blue!8}\textbf{43.7} & \cellcolor{blue!8}\textbf{71.5} \\
\midrule
\multirow{3}{*}{Qwen3-Coder}
  & \cellcolor{black!4}Original  & \cellcolor{black!4}512 & \cellcolor{black!4}{89.6} & \cellcolor{black!4}{80.0} & \cellcolor{black!4}85.4          & \cellcolor{black!4}{42.4} & \cellcolor{black!4}92.7          & \cellcolor{black!4}47.5          & \cellcolor{black!4}72.9          \\
  & REAP      & 384 & 87.5          & 70.0          & \textbf{86.4} & 37.9          & \textbf{94.5} & 47.7          & 70.7          \\
  & \cellcolor{blue!8}REAM      & \cellcolor{blue!8}384 & \cellcolor{blue!8}\textbf{89.3}          & \cellcolor{blue!8}\textbf{80.0} & \cellcolor{blue!8}85.3          & \cellcolor{blue!8}\textbf{40.4}          & \cellcolor{blue!8}\textbf{94.5} & \cellcolor{blue!8}\textbf{48.0} & \cellcolor{blue!8}\textbf{72.9} \\
\midrule
\multirow{3}{*}{GLM-4.5-Air}
  & \cellcolor{black!4}Original  & \cellcolor{black!4}128 & \cellcolor{black!4}90.4           & \cellcolor{black!4}83.3           & \cellcolor{black!4}94.8           & \cellcolor{black!4}42.9           & \cellcolor{black!4}93.9           & \cellcolor{black!4}57.4           & \cellcolor{black!4}77.1           \\
  & REAP      & 96  & 80.6           & 76.7           & 93.9           & \textbf{38.4}           & \textbf{90.2}           & 51.7           & 71.9           \\
  & \cellcolor{blue!8}REAM      & \cellcolor{blue!8}96  & \cellcolor{blue!8}\textbf{83.6}           & \cellcolor{blue!8}\textbf{83.3}           & \cellcolor{blue!8}\textbf{94.9}           & \cellcolor{blue!8}37.9           & \cellcolor{blue!8}\textbf{90.2}           & \cellcolor{blue!8}\textbf{53.7}           & \cellcolor{blue!8}\textbf{73.9}           \\
\bottomrule
\end{tabular}
\end{table}

\subsection{Additional Experiments}
 
\begin{figure}[t]
    \centering
\begin{subfigure}[t]{0.48\textwidth}
        \centering
        \includegraphics[width=0.82\textwidth]{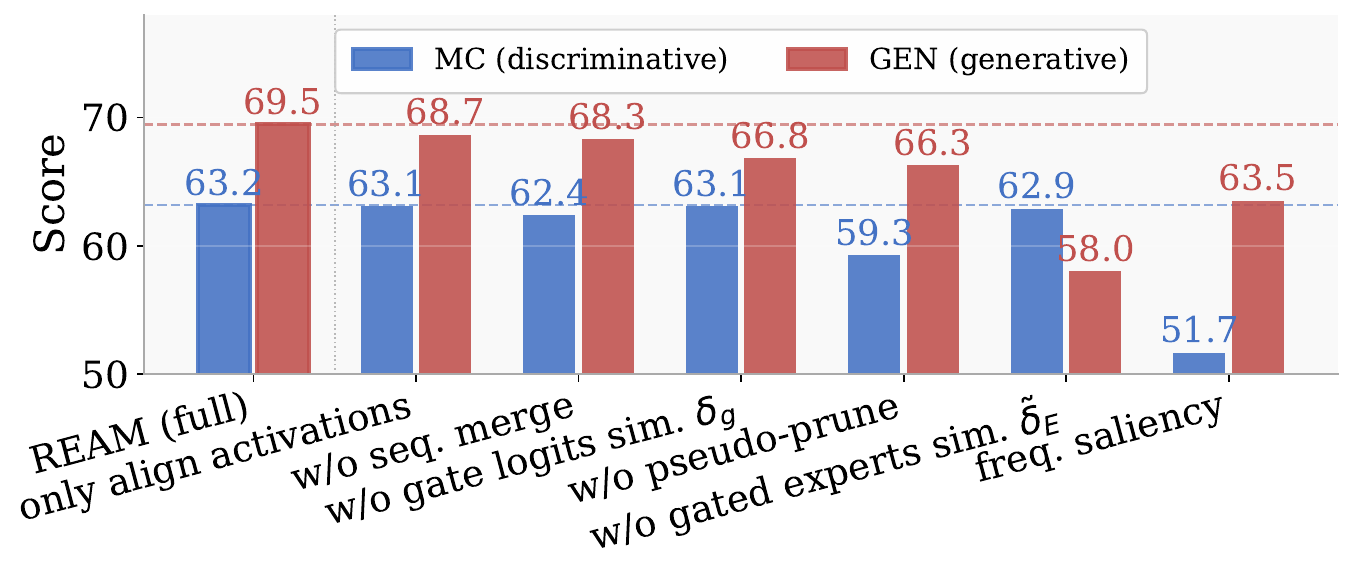}
        \caption{Avg. MC and GEN scores}
        \label{fig:ablation_avg_score}
            \end{subfigure}    
        \hfill
    \begin{subfigure}[t]{0.51\textwidth}
        \centering
        \includegraphics[width=\textwidth]{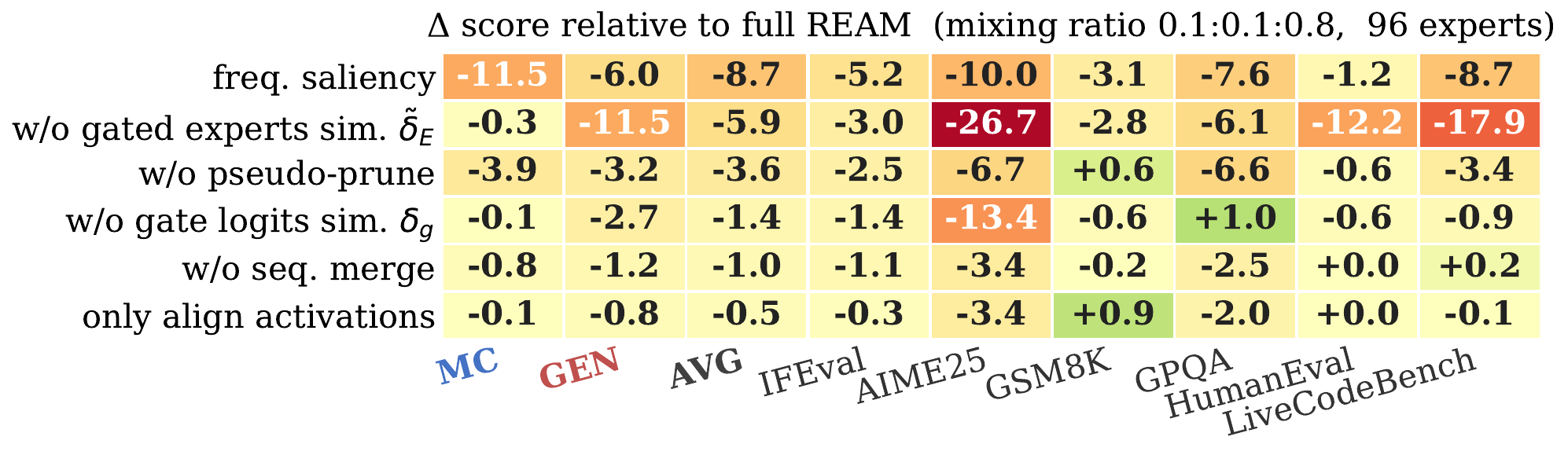}
        \caption{Per-task score drop ($\Delta$)}
        \label{fig:ablation_score_drop}
    \end{subfigure}
    \caption{\textbf{Ablation of REAM components}  with 96 experts:
           \textbf{(a)} \textcolor{blue}{MC} and \textcolor{red}{GEN} scores for each ablation variant;
           \textbf{(b)} Per-task score drop ($\Delta$) relative to the full REAM performance.}
    \label{fig:ablations}
    \vspace{-10pt}
\end{figure}

\paragraph{Ablation study.} Fig.~\ref{fig:ablation_avg_score} reports the effect of removing each REAM component in isolation at 96 experts with a GEN-favoring  calibration mixture of 0.1:0.1:0.8. 
We observe the largest single degradation ($\Delta$AVG = $-8.7$) to come from replacing REAP's saliency score (Eq. \eqref{eq:saliency_reap}) with routing frequency (Eq. \eqref{eq:saliency_freq}). This finding is in line with recent works confirming router frequency  as an unreliable proxy for expert importance given that it ignores the magnitude of each expert's actual contribution to the layer output \citep{lasby2025reap, mi2026effective}. Our second-largest drop stems from removing gate softmax scaling ($\sigma(\mathbf{x})$ in Eq. \eqref{eq:gated-expert-sim}) before computing pairwise output similarity ($\Delta$AVG = $-5.9$, $\Delta$GEN = $-11.5$) during grouping. This reaffirms that ignoring the router's confidence in grouping similarity treats all experts symmetrically, thus allowing experts that produce similar raw outputs but are preferred on different token distributions to be incorrectly merged. We also observe removing pseudo-pruning to incur a moderate penalty ($\Delta$AVG = $-3.6$), 
which confirms the importance of our grouping compared to the one used in MC-SMoE~\citep{limerge}.
We also find the expert co-activation signals from gate logit similarity ($\delta_g$ in Eq.~\eqref{eq:expert-sim}) and the re-computation of activations from sequential merging to be each contributing smaller but consistent gains of $\Delta$AVG = $-1.4$ and $-1.0$ respectively.  Finally, replacing the combined activation and weight alignment $\mathcal{C}_{\langle c_i, j\rangle}$ with activation-only alignment $\mathcal{C}_{\text{act}}$ yields the smallest penalty ($\Delta$AVG = $-0.5$), suggesting that the weight-based cost matrix  provides a marginal but consistent regularization in neuron pair matching (Section~\ref{sec:permutation}).
Removing all our components together would make REAM equivalent to MC-SMoE~\citep{limerge}.

\begin{wrapfigure}{r}{0.45\textwidth}
    \vspace{-1.2\baselineskip}
    \centering
    \includegraphics[width=0.40\textwidth]{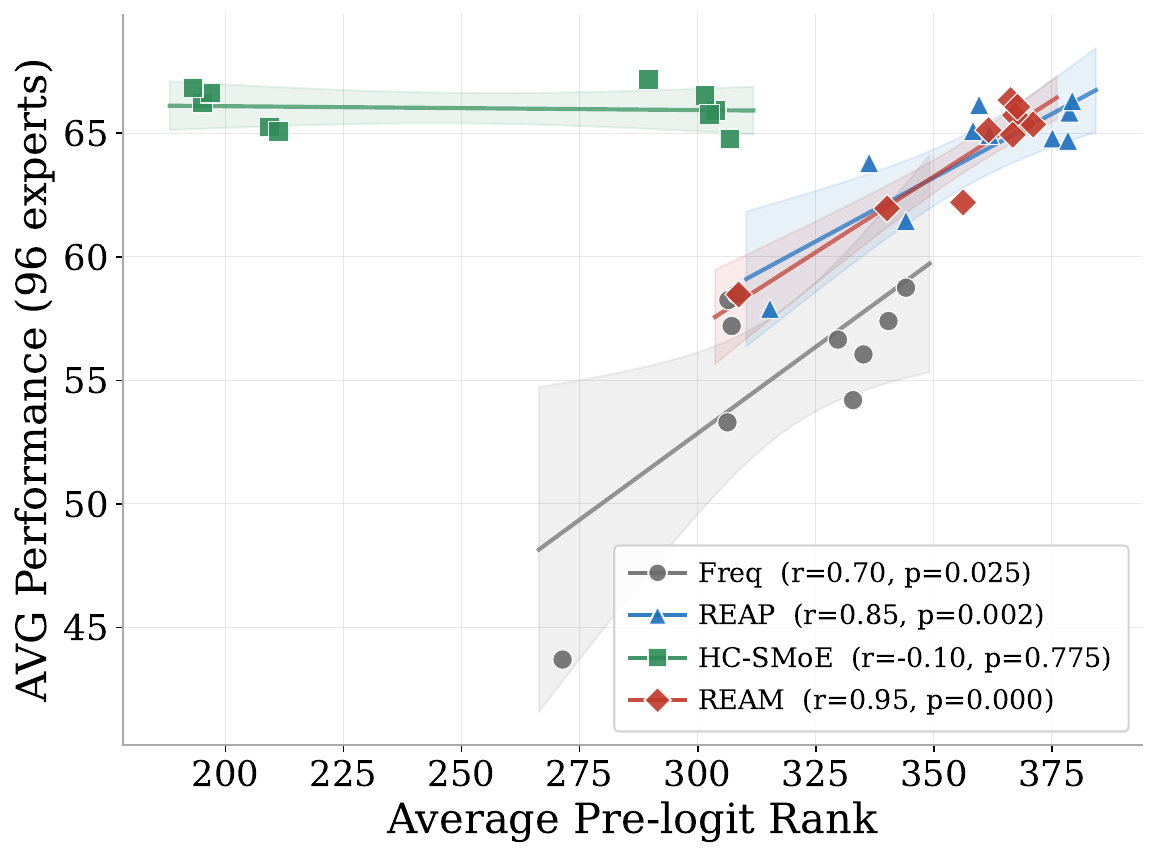}
    \caption{\textbf{Correlation} between avg. pre-logit ranks and AVG benchmark scores across 10 calibration ratios for 96 experts.}
    \label{fig:rank_correlations_96}
    \vspace{-1.0\baselineskip}
\end{wrapfigure}

\paragraph{Rank analyses.} To study whether expert merging strategies that better preserve the representational capacity of the compressed model  translate into higher benchmark scores, we compute the average numerical rank of the pre-logit embeddings for each method across all ten calibration mixtures and correlate it with the downstream performance. Fig. \ref{fig:rank_correlations_96} shows that REAM has the steepest and tightest regression curve where rank is an excellent predictor of performance. REAP follows closely but with a wider scatter while Freq has the least rank-efficiency. The strong correlation between rank and performances of these methods vouches for using rank as a cheap, task-agnostic proxy to estimate the optimal calibration mixtures in merging.
 
\section{Conclusion}

We propose REAM as an expert compression method that shows strong results across generative (GEN) benchmarks at 25\% and 50\% compression rates. We find several challenges for expert compression. First, no single method dominates across all setups and tasks: the baseline merging method (HC-SMoE) balances discriminative (MC) and generative (GEN) performance, while REAP and REAM can dominate either MC or GEN. Second, the trade-off between MC and GEN is surprising. MC tasks are generally considered easier, yet expert compression deteriorates them under certain calibration mixtures, indicating that MC and GEN  may rely on different subsets of experts. Understanding this asymmetry could inform mixture-aware compression methods that allocate capacity differently across expert groups. Finally, benchmarks with small sample sizes (e.g., AIME25 with 30 problems) introduce considerable variance, so future work should explore larger and more diverse evaluation suites to more accurately estimate the gap with the uncompressed models.\looseness-1

\section*{Acknowledgments}
Saurav Jha is supported by the IVADO postdoctoral fellowship and the Canada First Research Excellence Fund. The experiments were in part enabled by computational resources provided by Calcul Québec and Compute Canada. 

\bibliography{colm2026_conference}
\bibliographystyle{colm2026_conference}

\newpage
\appendix
\section{Appendix}

\subsection{Hyperparameters}
\label{app:hyperparam}
The only hyperparameter of REAM is group size $C$ of pseudo-pruning (Section~\ref{sec:pseudo-prune}) is fixed to 16 for Qwen3-30B-A3B-Instruct-2507 (when compressed to 96 experts) or to 32 (when compressed to 64 experts); 32 for Qwen3-Coder-Next and Qwen3-Next-80B-A3B-Instruct and to 16 for GLM-4.5-Air. A general idea behind this choice is that for models with more experts originally or more experts to be merged, we found it beneficial to increase $C$. This hyperparameter is not heavily tuned and is set once for each model and compression ratio.

\subsection{MC and GEN Tasks}
\label{app:mc_tasks}
The following 8 MC tasks are used for evaluation:
WinoGrande \citep{sakaguchi2019winogrande}, the Challenge and Easy set in AI2 Reasoning Challenge (ARC) \citep{clark2018think}, BoolQ \citep{clark2019boolq}, HellaSwag \citep{zellers2019hellaswag}, MMLU \citep{hendrycks2021measuring}, OpenBookQA \citep{mihaylov2018can}, and Recognizing Textual Entailment (RTE) \citep{bentivogli2009fifth}.
The following 6 generative tasks are used for evaluation: IFEval \citep{zhou2023instructionfollowingevaluationlargelanguage}, AIME25 \citep{aime25}, GSM8K \citep{cobbe2021gsm8k}, HumanEval \citep{chen2021evaluating}, GPQA-Diamond \citep{rein2024gpqa}, and LiveCodeBench-v6 \citep{jain2025livecodebench}. 

For evaluation we use EleutherAI Language Model Evaluation Harness~\citep{gao2021framework} with a HuggingFace or vLLM backend~\citep{kwon2023efficient} and default task settings. GPQA-Diamond is evaluated without chain-of-thought (CoT) reasoning using 5 shots. For LiveCodeBench-v6 we use their official evaluation code. But to evaluation GLM-4.5-Air on HumanEval and LiveCodeBench we use the evaluation tool from \url{https://github.com/zai-org/glm-simple-evals}.

\subsection{Why Evaluate on Different Mixtures of the Calibration Dataset?}
\label{subsec:calibration_ratios}
We note that expert merging is fundamentally a data-driven procedure, e.g., both the saliency scores ${S_i^{\text{reap}}}$ and the pairwise similarities ${\delta(c_i, j)}$ are computed entirely from activations on the calibration set $X$, thus making the merging decisions an implicit function of the calibration distribution. This has a direct consequence for downstream performance — if an expert is rarely activated or produces low-magnitude outputs on $X$, it will receive a low saliency score and be a candidate for absorption into another expert, regardless of how important it might be for a target task  underrepresented in $X$. Similarly, two experts that appear interchangeable on $X$ may serve very different roles on out-of-distribution inputs. The calibration set thus acts as an implicit prior over which expert behaviors to preserve. This motivates us to experiment extensively with different dataset mixtures (C4, Math, Code, and their combinations) to understand how compression quality varies with calibration distribution. In doing so, our hypothesis remains identifying a good compression method that does not remain tied to a fixed calibration assumption but  instead adapts its merging decisions flexibly to the target task distribution.

\begin{table}[h]
\centering
\caption{Calibration dataset mixing ratios used in experiments.
Each row defines the proportion of C4 (general text),
math (NuminaMath), and code (The-Stack-Smol).}
\small
\label{tab:ratios}
\begin{tabular}{c|c|c|c}
\hline
\textbf{C4} & \textbf{Math} & \textbf{Code} & \textbf{Description} \\
\hline
\hline
0.3 & 0.3 & 0.3 & Balanced \\
\hline
0.5 & 0.5 & 0.0 & C4 + math only \\
0.5 & 0.0 & 0.5 & C4 + code only \\
0.0 & 0.5 & 0.5 & Math + code only \\
\hline
0.2 & 0.5  & 0.3  & Math-leaning \\
0.1 & 0.8  & 0.1  & Math-heavy \\
0.0 & 0.7  & 0.3  & Math-heavy, no C4 \\
\hline
0.2 & 0.25 & 0.55 & Code-leaning \\
0.1 & 0.1  & 0.8  & Code-heavy \\
0.0 & 0.3  & 0.7  & Code-heavy, no C4 \\
\hline
\end{tabular}
\end{table}

\paragraph{Detailed analysis of calibration data vs. performance.}
\label{app:correlations_96_detailed}
We find C4 (general text) to be the strongest predictor of MC performance while The-Stack-Smol (code) to drive GEN performance (Fig.~\ref{fig:correlations_96}). Across Freq, REAP, and REAM, the proportion of C4 in the calibration mixture strongly predicts MC scores ($r \approx +0.95$--$+0.96$) while also  suppressing GEN scores ($r \approx -0.82$--$-0.85$). This can be attributed to MC benchmarks such as ARC, BoolQ, and HellaSwag drawing on the same factual and commonsense knowledge encoded in general web text. Subsequently, calibrating on C4 causes the saliency scores to favor the general-purpose experts that these tasks rely on but at the cost of the specialized experts that generative tasks require. Code data  shows a complementary pattern: positive correlation with GEN ($r \approx +0.59$--$+0.71$) and negative with MC ($r \approx -0.40$--$-0.57$), since code-heavy calibration elevates the saliency of structured-reasoning and syntax-specialized experts that directly serve GEN benchmarks like HumanEval and LiveCodeBench. Surprisingly, the proportion of math data has weak and near-zero correlations with both MC and GEN ($|r| \leq 0.19$ for REAP and REAM), despite AIME25 appearing in the GEN suite. This suggests that mathematical reasoning is distributed diffusely across experts rather than concentrated in a few high-activation specialists. As such, changing the math fraction does not systematically shift which experts survive merging. Put together, these findings suggest a fundamental MC--GEN trade-off. Because the merging budget is fixed, one cannot simultaneously preserve both general-text and code-specialized experts and the calibration data distribution acts as the sole lever for controlling this trade-off. Our REAM responds the best to this trade-off with its peak MC score of 69.2 at 96 experts (0.5:0.5:0) and its peak GEN score of 69.8 (0:0.5:0.5), beating all other methods on 96 experts. At 64 experts (50\% compression), REAM  achieves the best MC and the second-best GEN, maintaining a similar task-aligned pattern.

 \begin{figure}[t]
  \centering  \includegraphics[width=0.9\textwidth]{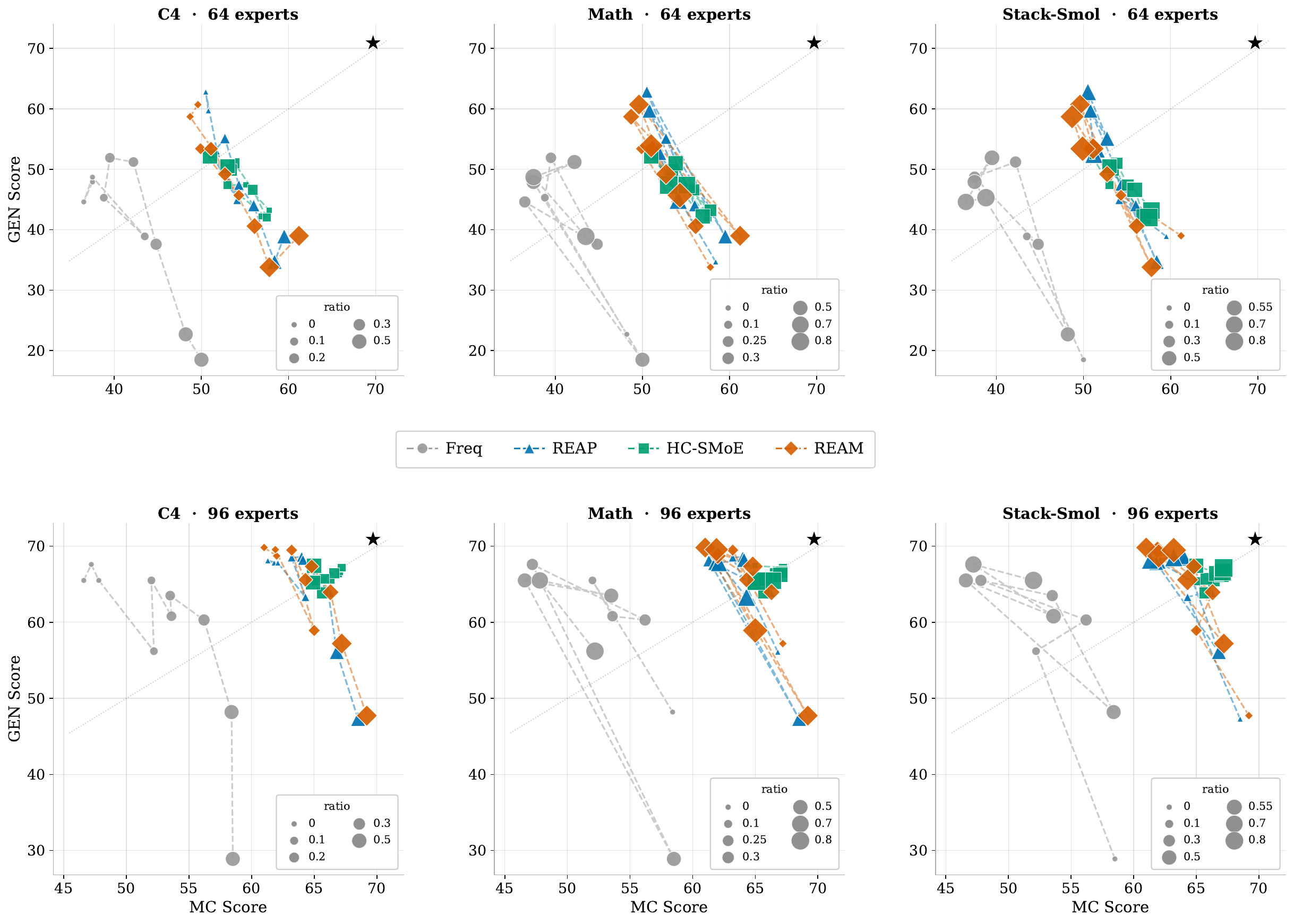}
  \caption{\textbf{Effect of calibration data mixture on MC–GEN trade-off.} Each panel shows discriminative (MC) vs.\ generative (GEN) benchmark scores for Freq, REAP, HC-SMoE, and REAM across ten mixing ratios of C4, Math, and Code datasets, with marker size proportional to each dataset's share of the mixture. Results are shown at two expert-count targets: 64 (50\% reduction) and 96 (25\% reduction). The star denotes the performance of the original Qwen3-30B-A3B-Instruct with 128 experts.}
  \label{fig:main_results_app}
\end{figure}

\subsection{Additional Ablations on Qwen3-Coder-Next}
Table \ref{tab:qwen3-coder-ablations} compares REAP against  various ablations of REAM components on a number of calibration mixtures. We see that the code-biased mixture 0.0:0.3:0.7 is the best overall for GEN average for all variants of REAM.
AIME25 is highly sensitive to the calibration mix, ranging from 53.3 (REAP, code-heavy 0.1:0.1:0.8) to 83.3 (w/o logit profile, 0.0:0.3:0.7), \textit{i.e.,} a $\sim$30-point swing.
Code-heavy calibration (0.1:0.1:0.8) also boosts GSM8K above the original (REAP: 89.7, REAM: 89.0 vs original 85.4), and pushes HumanEval to 95.1.  Both these results exceed the uncompressed model.
On the contrary, HumanEval is overall robust to compression where most variants stay in the 91–95 range regardless of method or ratio. We also find the removal of sequential merging to be the most damaging ablation where the performance for REAM at 0.0:0.3:0.7 ratio drops GEN from 72.9 to 69.0. Removing the logit profile similarity from pseudo-pruning surprisingly achieves the best AIME25 (83.3, which is even above the original 80.0) at 0.0:0.3:0.7. However,  this boost on AIME25 does not transfer to other ratios, suggesting that it may be a calibration interaction rather than a genuine gain.
Overall, we find  logit profile similarity to be the most important for maintaining a balanced GEN average across diverse ratios. 

\begin{table}[H]
\centering
\scriptsize
\setlength{\tabcolsep}{4pt}
\renewcommand{\arraystretch}{1.08}
\caption{Further GEN benchmark results for \textbf{Qwen3-Coder-Next} \citep{cao2026qwen3} compressed from $512$ to
         \textbf{384 experts (25\% reduction in $N$)}, with  group size $C=32$.
         The ratio column denotes the calibration mixture (C4\,:\,Math\,:\,Stack-Smol).
         \textbf{Bold} marks the best score in each column across all rows.}
\label{tab:qwen3-coder-ablations}
\begin{tabular}{l c| c c c c c c | c}
\toprule
\textbf{Method} & \textbf{Ratio} & \textbf{IFEval} & \textbf{AIME25}
  & \textbf{GSM8K} & \textbf{GPQA} & \textbf{HumanEval} & \textbf{LCB} & \textbf{GEN} \\
\midrule
\cellcolor{black!4}Original & \cellcolor{black!4}--- & \cellcolor{black!4}89.6 & \cellcolor{black!4}80.0 & \cellcolor{black!4}85.4 & \cellcolor{black!4}\textbf{42.4} & \cellcolor{black!4}92.7 & \cellcolor{black!4}47.5 & \cellcolor{black!4}\textbf{72.9} \\
\midrule
\multirow{4}{*}{REAP}
  & 0.0/0.3/0.7   & 87.5          & 70.0          & 86.4          & 37.9          & 94.5          & 47.7          & 70.7 \\
  & \cellcolor{black!4}0.1/0.1/0.8   & \cellcolor{black!4}87.5 & \cellcolor{black!4}53.3 & \cellcolor{black!4}\textbf{89.7} & \cellcolor{black!4}35.9 & \cellcolor{black!4}\textbf{95.1} & \cellcolor{black!4}47.6 & \cellcolor{black!4}68.2 \\
  & 0.2/0.25/0.55 & 86.6          & 60.0          & 87.6          & 37.9          & 93.3          & 47.0          & 68.7 \\
  & \cellcolor{black!4}0.2/0.5/0.3   & \cellcolor{black!4}88.1 & \cellcolor{black!4}60.0 & \cellcolor{black!4}86.1 & \cellcolor{black!4}34.3 & \cellcolor{black!4}89.6 & \cellcolor{black!4}42.7 & \cellcolor{black!4}66.8 \\
\midrule
\multirow{5}{*}{REAM full}
  & \cellcolor{blue!8}0.0/0.3/0.7   & \cellcolor{blue!8}89.3 & \cellcolor{blue!8}80.0 & \cellcolor{blue!8}85.3 & \cellcolor{blue!8}40.4 & \cellcolor{blue!8}94.5 & \cellcolor{blue!8}48.0 & \cellcolor{blue!8}\textbf{72.9} \\
  & \cellcolor{blue!8}0.1/0.1/0.8   & \cellcolor{blue!8}89.5 & \cellcolor{blue!8}60.0 & \cellcolor{blue!8}89.0 & \cellcolor{blue!8}36.4 & \cellcolor{blue!8}93.9 & \cellcolor{blue!8}44.0 & \cellcolor{blue!8}68.8 \\
  & \cellcolor{blue!8}0.2/0.25/0.55 & \cellcolor{blue!8}87.2 & \cellcolor{blue!8}60.0 & \cellcolor{blue!8}87.5 & \cellcolor{blue!8}36.9 & \cellcolor{blue!8}93.3 & \cellcolor{blue!8}41.0 & \cellcolor{blue!8}67.7 \\
  & \cellcolor{blue!8}0.0/0.7/0.3   & \cellcolor{blue!8}88.4 & \cellcolor{blue!8}56.7 & \cellcolor{blue!8}85.8 & \cellcolor{blue!8}38.9 & \cellcolor{blue!8}\textbf{95.1} & \cellcolor{blue!8}\textbf{48.7} & \cellcolor{blue!8}68.9 \\
  & \cellcolor{blue!8}0.0/0.5/0.5   & \cellcolor{blue!8}89.3 & \cellcolor{blue!8}73.3 & \cellcolor{blue!8}84.9 & \cellcolor{blue!8}39.4 & \cellcolor{blue!8}93.9 & \cellcolor{blue!8}48.4 & \cellcolor{blue!8}71.5 \\
\midrule
\multirow{3}{*}{REAM w/o $\delta_g$ in Eq.~\eqref{eq:expert-sim}}
  & 0.0/0.3/0.7   & \textbf{89.8} & \textbf{83.3} & 84.3          & 38.4          & 93.9          & 47.6          & \textbf{72.9} \\
  & \cellcolor{black!4}0.1/0.1/0.8   & \cellcolor{black!4}88.4 & \cellcolor{black!4}53.3 & \cellcolor{black!4}87.5 & \cellcolor{black!4}34.3 & \cellcolor{black!4}93.9 & \cellcolor{black!4}44.1 & \cellcolor{black!4}66.9 \\
  & 0.2/0.25/0.55 & 89.0          & 70.0          & 87.5          & 37.4          & 91.5          & 40.6          & 69.3 \\
\midrule
\multirow{3}{*}{REAM w/o seq.\ merge}
  & \cellcolor{black!4}0.0/0.3/0.7   & \cellcolor{black!4}89.3 & \cellcolor{black!4}63.3 & \cellcolor{black!4}84.6 & \cellcolor{black!4}38.4 & \cellcolor{black!4}92.1 & \cellcolor{black!4}46.4 & \cellcolor{black!4}69.0 \\
  & 0.1/0.1/0.8   & 88.4          & 63.3          & 87.9          & 31.3          & 93.3          & 43.6          & 68.0 \\
  & \cellcolor{black!4}0.2/0.25/0.55 & \cellcolor{black!4}89.1 & \cellcolor{black!4}70.0 & \cellcolor{black!4}87.0 & \cellcolor{black!4}36.9 & \cellcolor{black!4}93.3 & \cellcolor{black!4}41.9 & \cellcolor{black!4}69.7 \\
\bottomrule
\end{tabular}
\end{table}

\begin{table}[t]
\centering
\caption{Per-task generative (GEN) benchmark results on Qwen3-30B-A3B-Instruct-2507 \citep{yang2025qwen3} with 64 experts across all calibration mixing ratios, including one additional single-dataset REAM ratio. Columns show individual GEN tasks followed by aggregate MC, GEN, and overall averages. \textbf{Bold} marks the best result within each mixture-ratio block; \underline{underlined} marks the second best.}
\label{tab:gen_tasks_64}

\renewcommand{\arraystretch}{1.03}
\setlength{\tabcolsep}{2.6pt}
\scriptsize

\definecolor{headerbg}{RGB}{235,239,245}
\definecolor{subheaderbg}{RGB}{245,247,250}
\definecolor{baselinebg}{RGB}{244,244,244}
\definecolor{groupA}{RGB}{255,255,255}
\definecolor{groupB}{RGB}{248,249,251}
\definecolor{reambg}{RGB}{220,232,255}
\definecolor{headertext}{RGB}{0,0,0}

\begin{adjustbox}{max width=\textwidth,max totalheight=0.92\textheight,keepaspectratio}
\begin{tabular}{@{} l l c c c c c c c c c @{}}
\toprule
\rowcolor{headerbg}
\textcolor{headertext}{\textbf{Mix Ratio}} &
\textcolor{headertext}{\textbf{Method}} &
\textcolor{headertext}{\textbf{IFEval}} &
\textcolor{headertext}{\textbf{AIME25}} &
\textcolor{headertext}{\textbf{GSM8K}} &
\textcolor{headertext}{\textbf{GPQA}} &
\textcolor{headertext}{\textbf{HumanEval}} &
\textcolor{headertext}{\textbf{LiveCode}} &
\textcolor{headertext}{\textbf{MC}} &
\textcolor{headertext}{\textbf{GEN}} &
\textcolor{headertext}{\textbf{AVG}} \\
\cmidrule(lr){3-8}\cmidrule(lr){9-11}
\rowcolor{subheaderbg}
\textcolor{headertext}{\textbf{C4\,:\,Math\,:\,Code}} &
& & & & & & & & & \\
\midrule

\rowcolor{baselinebg}
\multicolumn{2}{@{}l}{\textit{Original} (128 experts)} &
\textit{90.4} & \textit{56.7} & \textit{89.3} & \textit{47.0} & \textit{93.3} & \textit{48.6} & \textit{69.7} & \textit{70.9} & \textit{70.3} \\
\midrule
\rowcolor{groupA}
\makecell[l]{0.3\,:\,0.3\,:\,0.3} & Freq    & 67.3 & 0.0 & 50.1 & 28.8 & \underline{66.5} & \underline{13.1} & 44.8 & 37.6 & 41.2 \\
\rowcolor{groupA}
                             & REAP    & \textbf{85.4} & \textbf{40.0} & \textbf{87.6} & \textbf{33.3} & 15.8 & 1.7 & \underline{56.0} & \underline{44.0} & \underline{50.0} \\
\rowcolor{groupA}
                             & HC-SMoE    & 77.2 & 23.3 & 70.9 & 20.7 & \textbf{79.3} & \textbf{28.2} & 53.4 & \textbf{49.9} & \textbf{51.7} \\
\rowcolor{reambg}
                             & REAM    & \underline{82.4} & \underline{33.3} & \underline{81.9} & \underline{31.3} & 13.4 & 1.0 & \textbf{56.1} & 40.5 & 48.3 \\
\cmidrule(lr){1-11}
\rowcolor{groupB}
\makecell[l]{0.1\,:\,0.8\,:\,0.1} & Freq    & 67.0 & 46.7 & 79.8 & \textbf{37.9} & 1.8 & 0.2 & 43.5 & 38.9 & 41.2 \\
\rowcolor{groupB}
                             & REAP    & \textbf{84.9} & \underline{50.0} & \underline{80.2} & \underline{35.4} & \underline{15.8} & \underline{3.4} & \underline{54.1} & 45.0 & 49.5 \\
\rowcolor{groupB}
                             & HC-SMoE    & 82.6 & 10.0 & 66.8 & \underline{35.4} & \textbf{68.9} & \textbf{20.6} & 53.0 & \textbf{47.4} & \textbf{50.2} \\
\rowcolor{reambg}
                             & REAM    & \underline{83.7} & \textbf{56.7} & \textbf{85.0} & \underline{35.4} & 11.6 & 2.0 & \textbf{54.3} & \underline{45.7} & \underline{50.0} \\
\cmidrule(lr){1-11}
\rowcolor{groupA}
\makecell[l]{0.5\,:\,0\,:\,0.5} & Freq    & 67.8 & \underline{0.0} & 1.8 & 25.8 & \underline{33.5} & \underline{7.1} & 48.2 & 22.7 & 35.4 \\
\rowcolor{groupA}
                             & REAP    & \textbf{82.4} & \underline{0.0} & \textbf{85.2} & \underline{32.3} & 7.9 & 0.6 & \textbf{58.4} & \underline{34.7} & \underline{46.6} \\
\rowcolor{groupA}
                             & HC-SMoE    & 73.4 & \textbf{23.3} & 70.2 & \textbf{33.8} & \textbf{75.0} & \textbf{27.5} & 53.0 & \textbf{50.5} & \textbf{51.8} \\
\rowcolor{reambg}
                             & REAM    & \underline{81.9} & \underline{0.0} & \underline{78.4} & 26.8 & 14.6 & 1.0 & \underline{57.8} & 33.8 & 45.8 \\
\cmidrule(lr){1-11}
\rowcolor{groupB}
\makecell[l]{0.5\,:\,0.5\,:\,0} & Freq    & 26.0 & 0.0 & 57.3 & 27.8 & 0.0 & \underline{0.0} & 50.0 & 18.5 & 34.3 \\
\rowcolor{groupB}
                             & REAP    & 77.5 & \textbf{36.7} & \underline{80.1} & \underline{33.8} & \underline{5.0} & \underline{0.0} & \underline{59.5} & 38.9 & 49.2 \\
\rowcolor{groupB}
                             & HC-SMoE    & \textbf{83.8} & 23.3 & 65.1 & 33.3 & \textbf{78.0} & \textbf{29.0} & 51.0 & \textbf{52.1} & \textbf{51.5} \\
\rowcolor{reambg}
                             & REAM    & \underline{77.9} & \underline{33.3} & \textbf{87.7} & \textbf{34.8} & 0.0 & \underline{0.0} & \textbf{61.2} & \underline{39.0} & \underline{50.1} \\
\cmidrule(lr){1-11}
\rowcolor{groupA}
\makecell[l]{0\,:\,0.5\,:\,0.5} & Freq    & 59.6 & 23.3 & 67.2 & 29.3 & \underline{77.4} & 30.4 & 37.5 & 47.9 & 42.7 \\
\rowcolor{groupA}
                             & REAP    & \textbf{86.0} & \underline{50.0} & \textbf{80.4} & \underline{33.3} & 76.8 & \underline{31.6} & \underline{50.8} & \underline{59.7} & \textbf{55.2} \\
\rowcolor{groupA}
                             & HC-SMoE    & 71.7 & 20.0 & 68.2 & \textbf{36.4} & 43.9 & 13.0 & \textbf{56.9} & 42.2 & 49.5 \\
\rowcolor{reambg}
                             & REAM    & \underline{80.2} & \textbf{60.0} & \underline{78.1} & 31.3 & \textbf{82.3} & \textbf{32.4} & 49.6 & \textbf{60.7} & \underline{55.2} \\
\cmidrule(lr){1-11}
\rowcolor{groupB}
\makecell[l]{0\,:\,0.3\,:\,0.7} & Freq    & 54.3 & 20.0 & 61.9 & 26.8 & 79.9 & 24.8 & 36.5 & 44.6 & 40.6 \\
\rowcolor{groupB}
                             & REAP    & \textbf{83.8} & \textbf{50.0} & \textbf{84.1} & \underline{31.8} & \textbf{89.0} & \textbf{38.3} & \underline{50.5} & \textbf{62.8} & \textbf{56.7} \\
\rowcolor{groupB}
                             & HC-SMoE    & 71.2 & 23.3 & 71.3 & \textbf{33.8} & 45.1 & 14.4 & \textbf{57.8} & 43.2 & 50.5 \\
\rowcolor{reambg}
                             & REAM    & \underline{79.5} & \underline{40.0} & \underline{82.0} & 28.8 & \underline{86.0} & \underline{35.8} & 48.7 & \underline{58.7} & \underline{53.7} \\
\cmidrule(lr){1-11}
\rowcolor{groupA}
\makecell[l]{0.1\,:\,0.1\,:\,0.8} & Freq    & 59.6 & 0.0 & 62.0 & \textbf{33.8} & \underline{82.9} & \textbf{33.2} & 38.8 & 45.2 & 42.0 \\
\rowcolor{groupA}
                             & REAP    & \textbf{83.9} & \textbf{26.7} & \textbf{86.7} & 25.8 & \textbf{90.2} & 1.7 & \underline{51.2} & \underline{52.5} & \textbf{51.9} \\
\rowcolor{groupA}
                             & HC-SMoE    & 67.9 & 20.0 & 71.2 & \underline{31.3} & 46.3 & 15.2 & \textbf{57.5} & 42.0 & 49.7 \\
\rowcolor{reambg}
                             & REAM    & \underline{78.5} & \textbf{26.7} & \underline{79.5} & 30.8 & 76.8 & \underline{27.8} & 49.9 & \textbf{53.4} & \underline{51.6} \\
\cmidrule(lr){1-11}
\rowcolor{groupB}
\makecell[l]{0\,:\,0.7\,:\,0.3} & Freq    & 62.5 & 33.3 & 66.0 & 34.3 & \textbf{78.0} & \underline{17.9} & 37.5 & 48.7 & 43.1 \\
\rowcolor{groupB}
                             & REAP    & \textbf{84.2} & \underline{46.7} & \underline{79.3} & 32.3 & 57.9 & 16.9 & \underline{51.8} & \underline{52.9} & \underline{52.3} \\
\rowcolor{groupB}
                             & HC-SMoE    & 77.3 & 20.0 & 68.4 & \textbf{36.4} & \underline{63.4} & \textbf{18.8} & \textbf{55.1} & 47.4 & 51.2 \\
\rowcolor{reambg}
                             & REAM    & \underline{79.9} & \textbf{50.0} & \textbf{81.0} & \underline{35.4} & 59.8 & 17.5 & 51.0 & \textbf{53.9} & \textbf{52.5} \\
\cmidrule(lr){1-11}
\rowcolor{groupA}
\makecell[l]{0.2\,:\,0.25\,:\,0.55} & Freq    & 68.2 & 20.0 & 77.0 & 27.8 & \textbf{84.2} & \textbf{34.1} & 39.5 & 51.9 & 45.7 \\
\rowcolor{groupA}
                             & REAP    & \textbf{88.1} & \textbf{41.0} & \textbf{86.7} & \underline{29.8} & 66.5 & 18.5 & \underline{52.7} & \textbf{55.1} & \textbf{53.9} \\
\rowcolor{groupA}
                             & HC-SMoE    & 72.4 & 20.0 & 75.4 & \textbf{33.3} & 58.5 & 20.1 & \textbf{55.9} & 46.6 & 51.3 \\
\rowcolor{reambg}
                             & REAM    & \underline{81.5} & \underline{33.3} & \underline{82.6} & 23.7 & \underline{74.4} & \underline{24.6} & 51.1 & \underline{53.4} & \underline{52.2} \\
\cmidrule(lr){1-11}
\rowcolor{groupB}
\makecell[l]{0.2\,:\,0.5\,:\,0.3} & Freq    & 71.7 & 36.7 & 73.8 & \underline{34.8} & \textbf{75.0} & \underline{15.0} & 42.2 & \textbf{51.2} & 46.7 \\
\rowcolor{groupB}
                             & REAP    & \textbf{84.7} & \underline{40.0} & \textbf{84.8} & 31.3 & 36.6 & 7.2 & \textbf{54.3} & 47.4 & 50.9 \\
\rowcolor{groupB}
                             & HC-SMoE    & 78.1 & 30.0 & 68.9 & \textbf{35.4} & \underline{70.1} & \textbf{23.4} & \underline{53.8} & \underline{51.0} & \textbf{52.4} \\
\rowcolor{reambg}
                             & REAM    & \underline{78.8} & \textbf{46.7} & \underline{82.9} & 32.8 & 45.7 & 8.5 & 52.7 & 49.2 & \underline{51.0} \\
\cmidrule(lr){1-11}
\rowcolor{reambg}
\makecell[l]{1\,:\,0\,:\,0} & REAM    & 74.3 & 0.0 & 73.6 & 26.8 & 0.0 & 0.0 & 64.7 & 29.1 & 46.9 \\

\bottomrule
\end{tabular}
\end{adjustbox}
\end{table}

\begin{table}[t]
\centering
\caption{Per-task generative (GEN) benchmark results on Qwen3-30B-A3B-Instruct-2507 \citep{yang2025qwen3} with 96 experts across all calibration mixing ratios, including three additional single-dataset REAM ratios. Columns show individual GEN tasks followed by aggregate MC, GEN, and overall averages. \textbf{Bold} marks the best result within each mixture-ratio block; \underline{underlined} marks the second best.}
\label{tab:gen_tasks_96}

\renewcommand{\arraystretch}{1.03}
\setlength{\tabcolsep}{2.6pt}
\scriptsize

\definecolor{headerbg}{RGB}{235,239,245}
\definecolor{subheaderbg}{RGB}{245,247,250}
\definecolor{baselinebg}{RGB}{244,244,244}
\definecolor{groupA}{RGB}{255,255,255}
\definecolor{groupB}{RGB}{248,249,251}
\definecolor{reambg}{RGB}{220,232,255}
\definecolor{headertext}{RGB}{0,0,0}

\begin{adjustbox}{max width=\textwidth,max totalheight=0.92\textheight,keepaspectratio}
\begin{tabular}{@{} l l c c c c c c c c c @{}}
\toprule
\rowcolor{headerbg}
\textcolor{headertext}{\textbf{Mix Ratio}} &
\textcolor{headertext}{\textbf{Method}} &
\textcolor{headertext}{\textbf{IFEval}} &
\textcolor{headertext}{\textbf{AIME25}} &
\textcolor{headertext}{\textbf{GSM8K}} &
\textcolor{headertext}{\textbf{GPQA}} &
\textcolor{headertext}{\textbf{HumanEval}} &
\textcolor{headertext}{\textbf{LiveCode}} &
\textcolor{headertext}{\textbf{MC}} &
\textcolor{headertext}{\textbf{GEN}} &
\textcolor{headertext}{\textbf{AVG}} \\
\cmidrule(lr){3-8}\cmidrule(lr){9-11}
\rowcolor{subheaderbg}
\textcolor{headertext}{\textbf{C4\,:\,Math\,:\,Code}} &
& & & & & & & & & \\
\midrule

\rowcolor{baselinebg}
\multicolumn{2}{@{}l}{\textit{Original} (128 experts)} &
\textit{90.4} & \textit{56.7} & \textit{89.3} & \textit{47.0} & \textit{93.3} & \textit{48.6} & \textit{69.7} & \textit{70.9} & \textit{70.3} \\
\midrule
\rowcolor{groupA}
\makecell[l]{0.3\,:\,0.3\,:\,0.3} & Freq    & 84.0 & \underline{43.3} & 83.3 & 31.8 & 80.5 & \underline{39.0} & 56.2 & 60.3 & 58.3 \\
\rowcolor{groupA}
                             & REAP    & \textbf{89.2} & \textbf{63.3} & \underline{86.1} & \textbf{40.4} & 75.6 & 30.1 & \underline{66.1} & \textbf{64.1} & \underline{65.1} \\
\rowcolor{groupA}
                             & HC-SMoE    & 88.4 & 40.0 & 84.2 & 34.3 & \textbf{91.5} & \textbf{44.7} & 65.7 & 63.9 & 64.8 \\
\rowcolor{reambg}
                             & REAM    & \underline{88.7} & \underline{43.3} & \textbf{87.3} & \underline{39.4} & \underline{88.4} & 36.6 & \textbf{66.3} & \underline{64.0} & \textbf{65.1} \\
\cmidrule(lr){1-11}
\rowcolor{groupB}
\makecell[l]{0.1\,:\,0.8\,:\,0.1} & Freq    & 87.3 & \textbf{60.0} & 84.9 & 35.4 & 54.9 & 15.0 & 52.2 & 56.2 & 54.2 \\
\rowcolor{groupB}
                             & REAP    & \underline{88.4} & \textbf{60.0} & \underline{85.1} & \textbf{38.9} & \underline{77.4} & \underline{29.9} & 64.3 & \underline{63.3} & \underline{63.8} \\
\rowcolor{groupB}
                             & HC-SMoE    & \textbf{89.7} & 46.7 & 85.0 & \underline{36.9} & \textbf{91.5} & \textbf{42.6} & \textbf{65.1} & \textbf{65.4} & \textbf{65.2} \\
\rowcolor{reambg}
                             & REAM    & 88.0 & 40.0 & \textbf{88.8} & 35.4 & 75.0 & 26.3 & \underline{65.0} & 58.9 & 62.0 \\
\cmidrule(lr){1-11}
\rowcolor{groupA}
\makecell[l]{0.5\,:\,0\,:\,0.5} & Freq    & 83.2 & 0.0 & 68.5 & 32.8 & 73.8 & 30.9 & 58.4 & 48.2 & 53.3 \\
\rowcolor{groupA}
                             & REAP    & \textbf{89.7} & \underline{13.3} & \textbf{86.8} & \underline{35.9} & 81.7 & 29.3 & \underline{66.8} & 56.1 & 61.5 \\
\rowcolor{groupA}
                             & HC-SMoE    & 88.2 & \textbf{60.0} & 84.7 & 34.3 & \textbf{91.5} & \textbf{45.9} & 65.0 & \textbf{67.4} & \textbf{66.2} \\
\rowcolor{reambg}
                             & REAM    & \underline{89.0} & \underline{13.3} & \underline{85.9} & \textbf{36.4} & \underline{85.4} & \underline{33.2} & \textbf{67.2} & \underline{57.2} & \underline{62.2} \\
\cmidrule(lr){1-11}
\rowcolor{groupB}
\makecell[l]{0.5\,:\,0.5\,:\,0} & Freq    & 56.1 & 10.0 & 71.1 & 35.4 & 0.6 & 0.0 & 58.5 & 28.9 & 43.7 \\
\rowcolor{groupB}
                             & REAP    & 88.2 & \textbf{66.7} & \underline{85.7} & \textbf{40.4} & \underline{2.4} & \underline{0.2} & \underline{68.5} & 47.3 & 57.9 \\
\rowcolor{groupB}
                             & HC-SMoE    & \underline{89.3} & 43.3 & 84.9 & 36.4 & \textbf{92.1} & \textbf{45.4} & 64.9 & \textbf{65.2} & \textbf{65.1} \\
\rowcolor{reambg}
                             & REAM    & \textbf{89.6} & \textbf{66.7} & \textbf{87.2} & \textbf{40.4} & \underline{2.4} & 0.1 & \textbf{69.2} & \underline{47.7} & \underline{58.5} \\
\cmidrule(lr){1-11}
\rowcolor{groupA}
\makecell[l]{0\,:\,0.5\,:\,0.5} & Freq    & 86.3 & 50.0 & 79.6 & 32.3 & \textbf{94.5} & \underline{50.0} & 46.6 & 65.5 & 56.0 \\
\rowcolor{groupA}
                             & REAP    & 88.4 & \underline{56.7} & 84.9 & \textbf{38.4} & 91.5 & 46.8 & \underline{61.8} & \underline{67.8} & 64.8 \\
\rowcolor{groupA}
                             & HC-SMoE    & \underline{88.8} & 53.3 & \underline{85.0} & 36.4 & 91.5 & 42.5 & \textbf{67.0} & 66.2 & \textbf{66.6} \\
\rowcolor{reambg}
                             & REAM    & \textbf{89.9} & \textbf{60.0} & \textbf{86.3} & \textbf{38.4} & \underline{93.3} & \textbf{51.0} & 61.0 & \textbf{69.8} & \underline{65.4} \\
\cmidrule(lr){1-11}
\rowcolor{groupB}
\makecell[l]{0\,:\,0.3\,:\,0.7} & Freq    & 87.8 & \textbf{60.0} & 82.9 & 36.9 & \textbf{93.9} & 44.0 & 47.2 & 67.6 & 57.4 \\
\rowcolor{groupB}
                             & REAP    & \underline{89.1} & 50.0 & \underline{87.3} & \textbf{42.4} & \underline{92.7} & \underline{47.0} & 61.3 & \underline{68.1} & 64.7 \\
\rowcolor{groupB}
                             & HC-SMoE    & 87.4 & \underline{56.7} & 85.3 & 36.4 & 90.2 & 43.5 & \textbf{67.1} & 66.6 & \textbf{66.8} \\
\rowcolor{reambg}
                             & REAM    & \textbf{90.9} & 53.3 & \textbf{87.7} & \underline{40.9} & 91.5 & \textbf{48.0} & \underline{62.0} & \textbf{68.7} & \underline{65.4} \\
\cmidrule(lr){1-11}
\rowcolor{groupA}
\makecell[l]{0.1\,:\,0.1\,:\,0.8} & Freq    & 83.0 & 46.7 & \textbf{88.8} & 36.9 & 87.8 & \underline{49.9} & 52.0 & 65.5 & 58.8 \\
\rowcolor{groupA}
                             & REAP    & \underline{89.2} & \textbf{56.7} & 85.1 & 37.4 & \textbf{92.7} & \textbf{50.1} & \underline{63.2} & \underline{68.5} & 65.9 \\
\rowcolor{groupA}
                             & HC-SMoE    & 88.0 & \textbf{56.7} & 85.8 & \underline{38.4} & 91.5 & 42.6 & \textbf{67.2} & 67.2 & \textbf{67.2} \\
\rowcolor{reambg}
                             & REAM    & \textbf{91.7} & \textbf{56.7} & \underline{87.6} & \textbf{38.9} & \textbf{92.7} & 49.3 & \underline{63.2} & \textbf{69.5} & \underline{66.3} \\
\cmidrule(lr){1-11}
\rowcolor{groupB}
\makecell[l]{0\,:\,0.7\,:\,0.3} & Freq    & 87.2 & 53.3 & 79.1 & 34.8 & \textbf{92.7} & \underline{45.8} & 47.8 & 65.5 & 56.6 \\
\rowcolor{groupB}
                             & REAP    & 87.6 & \underline{60.0} & \underline{84.8} & \textbf{37.9} & \underline{91.5} & 45.0 & \underline{62.1} & \underline{67.8} & 65.0 \\
\rowcolor{groupB}
                             & HC-SMoE    & \textbf{89.6} & 50.0 & 83.9 & 35.9 & 90.2 & 43.1 & \textbf{66.4} & 65.5 & \textbf{65.9} \\
\rowcolor{reambg}
                             & REAM    & \underline{89.0} & \textbf{63.3} & \textbf{86.8} & \underline{36.9} & 90.8 & \textbf{50.5} & 61.9 & \textbf{69.5} & \underline{65.7} \\
\cmidrule(lr){1-11}
\rowcolor{groupA}
\makecell[l]{0.2\,:\,0.25\,:\,0.55} & Freq    & 83.5 & 30.0 & 81.3 & 32.8 & 87.8 & \underline{49.4} & 53.6 & 60.8 & 57.2 \\
\rowcolor{groupA}
                             & REAP    & 89.6 & \textbf{50.0} & \textbf{87.9} & \textbf{39.4} & \textbf{94.5} & \textbf{50.3} & 64.0 & \textbf{68.6} & \underline{66.3} \\
\rowcolor{groupA}
                             & HC-SMoE    & \underline{89.8} & \textbf{50.0} & 84.4 & \underline{38.9} & 91.5 & 44.0 & \textbf{66.6} & \underline{66.4} & \textbf{66.5} \\
\rowcolor{reambg}
                             & REAM    & \textbf{90.3} & 43.3 & \underline{87.6} & 33.8 & \textbf{94.5} & 44.0 & \underline{64.3} & 65.6 & 64.9 \\
\cmidrule(lr){1-11}
\rowcolor{groupB}
\makecell[l]{0.2\,:\,0.5\,:\,0.3} & Freq    & 82.1 & 50.0 & 83.0 & \underline{35.4} & 85.4 & \underline{45.2} & 53.5 & 63.5 & 58.5 \\
\rowcolor{groupB}
                             & REAP    & \textbf{89.3} & \textbf{63.3} & \underline{85.4} & \textbf{39.9} & 86.6 & 44.6 & 64.1 & \textbf{68.2} & \textbf{66.1} \\
\rowcolor{groupB}
                             & HC-SMoE    & \textbf{89.3} & 53.3 & 84.8 & 34.3 & \underline{89.6} & 42.9 & \textbf{65.8} & 65.7 & 65.8 \\
\rowcolor{reambg}
                             & REAM    & 88.0 & \underline{56.7} & \textbf{88.4} & \underline{35.4} & \textbf{90.2} & \textbf{45.3} & \underline{64.8} & \underline{67.3} & \underline{66.1} \\
\cmidrule(lr){1-11}
\rowcolor{reambg}
\makecell[l]{0\,:\,1\,:\,0} & REAM    & 88.8 & 56.7 & 87.6 & 35.9 & 71.3 & 28.5 & 64.3 & 61.5 & 62.9 \\
\rowcolor{reambg}
\makecell[l]{0\,:\,0\,:\,1} & REAM    & 92.2 & 60.0 & 88.0 & 32.8 & 92.7 & 49.3 & 62.9 & 69.2 & 66.0 \\
\rowcolor{reambg}
\makecell[l]{1\,:\,0\,:\,0} & REAM    & 87.9 & 0.0 & 87.0 & 37.9 & 0.0 & 0.0 & 69.6 & 35.5 & 52.5 \\

\bottomrule
\end{tabular}
\end{adjustbox}
\end{table}

\begin{figure}[hbp]
\vspace{-15pt}
    \centering
   \includegraphics[width=0.40\textwidth]{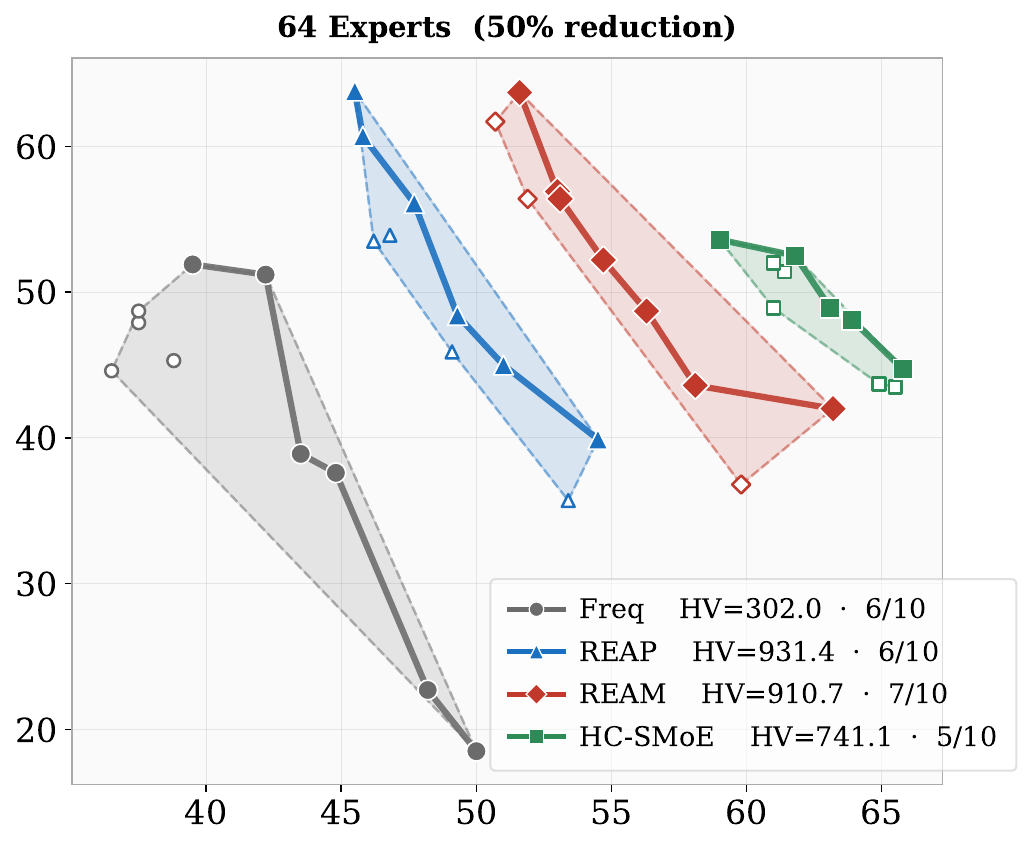}
    \caption{\textbf{Pareto frontiers of expert-merging methods at 64 retained experts.} Each point is one of 10 calibration mixtures; filled markers denote Pareto-optimal configurations (not simultaneously dominated on both MC and GEN by any other mixture of the same method) and hollow markers denote dominated ones. The hypervolume (HV) measures the area of the MC$\times$GEN plane dominated by each method's frontier relative to a shared reference point, quantifying its overall performance ceiling. HV and $n/10$ counts are computed on the original scores. Per-method offsets are then applied for better visibility.}
    \label{fig:pareto_64}
    \vspace{-15pt}
\end{figure}

\end{document}